\documentclass{article}
\usepackage{iclr2027_conference,times}
\usepackage{amsmath,amssymb}
\usepackage{booktabs,array,graphicx}
\usepackage{colortbl}
\usepackage{float}
\usepackage{needspace}
\usepackage{algorithm}
\usepackage{algpseudocode}
\usepackage{enumitem}
\usepackage{wrapfig}
\usepackage{caption}

\usepackage{url,xcolor}

\definecolor{citecyan}{HTML}{007C91}
\definecolor{citebrown}{HTML}{8B5A2B}
\definecolor{citeNavy}{HTML}{3C5488}     
\definecolor{citeBrown}{HTML}{7E6148}    
\definecolor{citePurple}{HTML}{6A5F7D}   
\definecolor{citecolor}{HTML}{0071BC}
\definecolor{citeFashion}{HTML}{BF8040}

\usepackage[
  colorlinks=true,
  linkcolor=red,
  citecolor=citecyan,
  urlcolor=black
]{hyperref}
\hypersetup{
  pdftitle={Are We Evaluating VLM Token Pruning Right? Looking Beyond Accuracy and Token Count},
  pdfauthor={Geng Li, Guohao Chen, Shilin Shan, Jianfei Yang}
}

\definecolor{draftblue}{RGB}{35,90,160}

\newcommand{\Dplus}{\mathcal{D}^{+}}
\newcommand{\Dminus}{\mathcal{D}^{-}}
\newcommand{\ind}{\mathbf{1}}

\newcommand{\PCS}{\mathrm{PCS}}
\newtheorem{proposition}{Proposition}
\newcolumntype{L}[1]{>{\raggedright\arraybackslash}p{#1}}

\title{Are We Evaluating VLM Token Pruning Right?\\Looking Beyond Accuracy and Token Count}
\author{%
\begin{minipage}{0.95\textwidth}
\vspace*{6pt}
\raggedright
{\large\bfseries
Geng Li$^{1}$\thanks{Equal contribution.}~~,~
Guohao Chen$^{1*}$~~,~
Shilin Shan$^{1}$~~,~
Jianfei Yang$^{1}$}\\[2pt]
{\normalsize\normalfont Nanyang Technological University$^{1}$}
\end{minipage}}
\iclrfinalcopy

\begin{document}
\maketitle
\lhead{Preprint}
\vspace{-0.18in}

\begin{abstract}
Visual token pruning aims to reduce inference cost while preserving the capabilities of the original VLM. Yet existing evaluations rely on aggregate task accuracy and retained token count as proxies for pruning quality and actual efficiency.\linebreak We reveal two limitations: 1) Aggregate accuracy can obscure whether a pruning method truly identifies critical visual tokens—many examples remain solvable under arbitrary token removal and models can exploit textual shortcuts when visual evidence is degraded. 2) Retained token count is an unreliable proxy for efficiency, as identical token budgets can incur different end-to-end costs depending on where and how pruning is performed. Consequently, comparisons based on aggregate accuracy and retained token counts fail to recover the global performance–efficiency Pareto frontier across methods. We introduce PAA-Bench, a \textit{\textbf{plug-and-play}} evaluation framework for existing benchmarks, to evaluate whether pruning succeeds through meaningful token selection and whether this success translates into real speedup. To determine when token choice actually matters, we propose \textit{Random-Pass@\(k\)} to measure how often an example remains correct under random pruning, enabling comprehensive evaluation across different levels of selection difficulty. To compare efficiency fairly, we introduce \textit{Preservation-Constrained Speedup} (PCS) to report the maximum end-to-end speedup each method achieves under\linebreak a shared preservation target. Experiments across \textit{8} methods, \textit{3} VLM backbones, and \textit{9} benchmarks show that stronger token compression does not necessarily yield greater speedup, method differences widen on examples where token selection matters, and efficiency rankings change with the preservation target.
\end{abstract}

\suppressfloats[t]
\begingroup
\setlength{\intextsep}{6pt}
\setlength{\abovecaptionskip}{5pt}
\begin{figure}[t]
    \vspace{-10pt}
    \centering
    \includegraphics[width=0.97\linewidth]{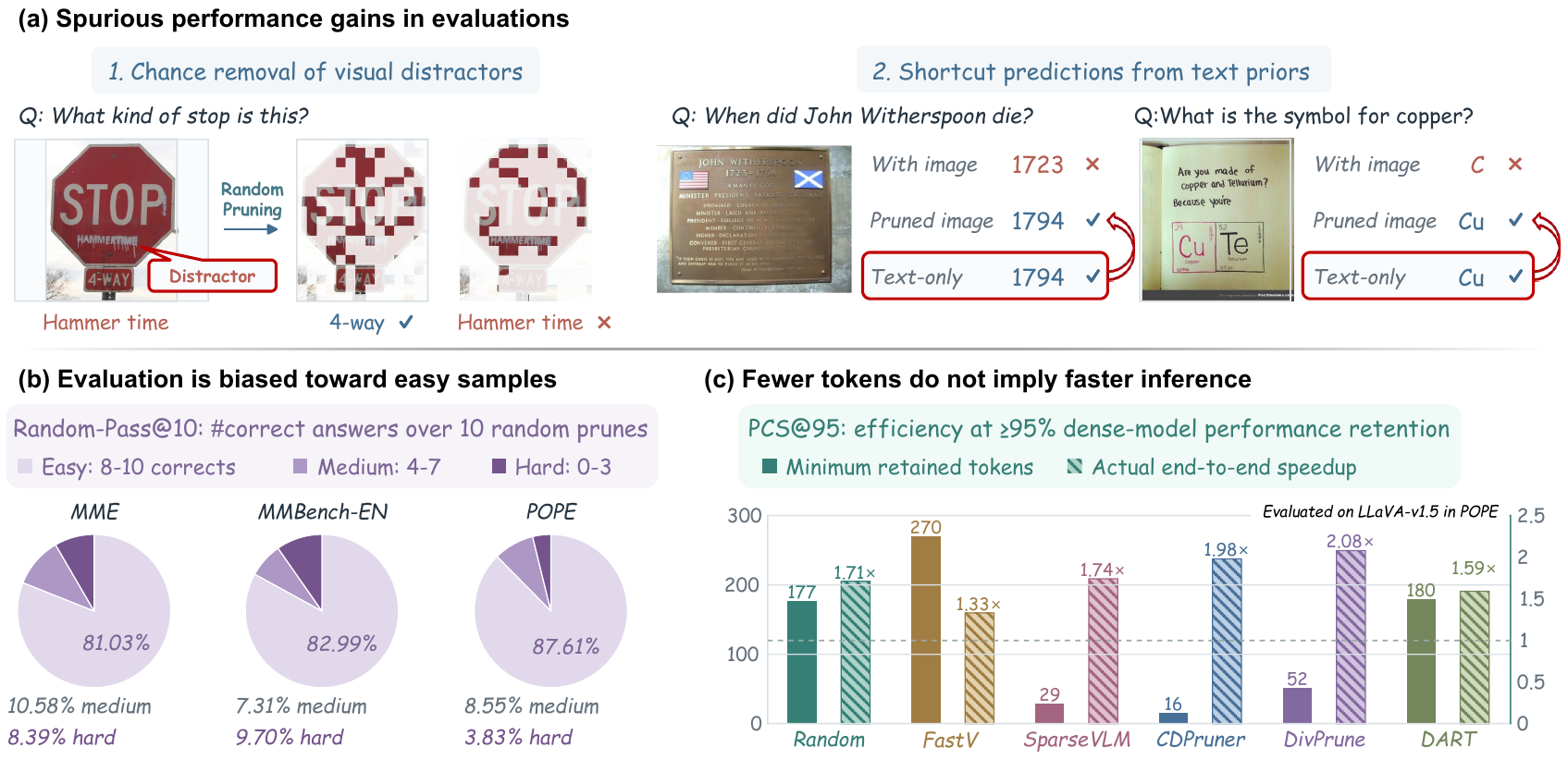}
    \caption{\textbf{Pitfalls in visual-token pruning evaluation.}
    (a) Corrections can arise from distractor removal or text priors and mask losses of originally correct predictions.
    (b) Under the budget of 64 tokens, easy examples dominate MME, MMBench-EN and POPE under Random-Pass@10.
    (c) At 95\% correctness preservation, fewer retained tokens do not guarantee greater end-to-end speedup.}
    \label{fig:evaluation_gaps}
\end{figure}
\vspace{-18pt}
\endgroup

\section{Introduction}
\label{sec:introduction}

Vision-language models (VLMs) are increasingly designed to process large and diverse visual inputs, including high-resolution images, multiple images, and videos~\citep{lin2024videollava,wang2024qwen2}. Supporting such inputs naturally requires longer visual-token sequences, increasing both computation and cache costs at inference~\citep{su2026gptailor}. This growth in token length makes efficient inference a key challenge for VLM deployment. However, enhancing efficiency through retraining requires substantial additional computation and data, and needs to be repeated when given new models. This makes training-free acceleration particularly attractive, as it can be applied directly to pretrained VLMs without model-specific retraining~\citep{men2025shortgpt,zhang2025sparsevlm}.

Among training-free acceleration approaches, visual token pruning~\citep{takezoe2026learnpruner} directly shortens the visual sequence passed to later layers, avoiding computation on redundant visual information~\citep{alvar2025divprune,zhang2026beyond}. An effective pruning method should therefore translate this token reduction into actual inference acceleration while retaining the predictive capability of the original VLM. Accordingly, existing studies typically vary the number of visual tokens retained after pruning and report how model accuracy and inference time change as more tokens are removed~\citep{yang2025visionzip}. Recent studies, such as G-Prune~\citep{jiang2025gprune} and FEATHER \citep{endo2025feather}, have also illustrated that pruning sensitivity varies across tasks and benchmarks.

Despite these efforts, as illustrated in Figure~\ref{fig:evaluation_gaps}, current evaluation can still provide an incomplete picture of pruning quality and realized efficiency. \textbf{\textit{(i) Many benchmark examples do not truly test token-selection ability.}} We find that, across commonly used benchmarks such as GQA~\citep{hudson2019gqa}, MMBench~\citep{liu2024mmbench}, and POPE~\citep{li2023evaluating}, more than 80\% of examples remain correct in 8--10 out of 10 \textit{random-pruning} trials under most evaluated settings. These pruning-insensitive examples can dominate aggregate results and obscure meaningful differences among token-selection methods. \textbf{\textit{(ii) Aggregate accuracy does not directly measure capability preservation.}} Final accuracy mixes pruning-induced failures with wrong-to-correct transitions, which can arise from distractor removal or increased reliance on textual priors rather than better preservation of visual information. Such spurious performance gains can offset newly introduced errors and mask the loss of capabilities in the original VLM. \textbf{\textit{(iii) Retained token count is an unreliable proxy for realized efficiency.}} Methods retaining the same number of tokens can incur substantially different end-to-end costs because pruning location and algorithmic overhead differ across methods. Together, these limitations show that token pruning evaluation should answer three distinct questions: ``\textit{whether performance depends on meaningful token selection, whether the model’s capability is preserved, and how much realized speedup is achieved under that preservation level}".

To answer these questions, we introduce \textbf{PAA-Bench}, a \textit{plug-and-play evaluation framework} for existing benchmarks that separates token-selection effectiveness, correctness preservation, and realized efficiency. Our central idea is to replace proxy-based comparison with direct measurements of whether token choice matters, what capability is preserved, and how much real speedup is obtained at that preservation level. Accordingly, we propose \textbf{\textit{Random-Pass@}$k$}, which uses repeated random pruning as a method-independent reference to determine whether successful predictions genuinely depend on token choice. To assess what pruning actually preserves, we then define performance rigorously by preservation of the dense model's correct predictions, so that spurious correction gains are not mistaken for preserved capability. Using capability preservation as the performance criterion, we introduce \textit{\textbf{Preservation-Constrained Speedup}} (PCS), which reports each method's maximum measured end-to-end speedup under the same preservation target, including pruning overhead. 

PAA-Bench therefore \textbf{shifts the basis of comparison from equal token counts to equal capability preservation, while separately revealing whether success actually depends on token selection}:


\textbf{Main contributions.} \textit{(1)} We rethink conventional token-pruning evaluation and reveal that comparisons based on aggregate accuracy and retained token count can obscure whether success stems from meaningful token selection, whether the original model's capability is preserved, and whether compression translates into realized efficiency. \textit{(2)} We introduce \textbf{PAA-Bench}, a plug-and-play framework for existing benchmarks that evaluates token-selection effectiveness, correctness preservation, and realized efficiency through Random-Pass@$k$, dense-model correctness preservation, and Preservation-Constrained Speedup, with Random-Pass@$k$ theoretically characterizing the headroom available to informed selection from random-pruning success. \textit{(3)} Across 8 pruning methods, 3 VLM backbones, and 9 benchmarks, PAA-Bench reveals a more discriminative view of the performance–efficiency trade-offs: preservation-aware scoring can reverse rankings, selection-sensitive examples expose larger method gaps, and efficiency rankings vary by preservation level.

\section{Related Work}
\label{sec:related_work}

\begin{figure}[!t]
    \centering
    \includegraphics[width=\linewidth]{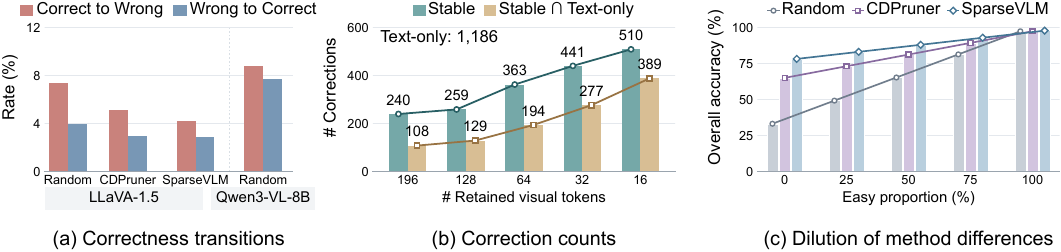}
  \caption{\textbf{Diagnostics of token-pruning evaluation.}
(a) \textbf{Correction gains can mask preservation losses}: correct-to-wrong
and wrong-to-correct rates for LLaVA-1.5 at 128 tokens (left) and Qwen3-VL-8B
at 20\% token retention (right).
(b) \textbf{Stable corrections overlap with text-only behavior}: counts of
Stable, Text-only corrections, and their intersection across budgets on LLaVA-1.5.
(c) \textbf{Method gaps widen as the Easy-example
proportion decreases}: overall accuracy of Random, CDPruner, and SparseVLM
across mixtures with 0--100\% Easy examples. Details in Appendix~\ref{app:dataset-difficulty-distributions}.}
\label{fig:evaluation_diagnostics}
\end{figure}

\textbf{Visual token reduction.}
Existing visual-token pruning methods~\citep{li2025tokenpacker,Vasu_2025_CVPR,Xing_2025_CVPR,huang2025dynamic,zhang2025llava} use different signals to prune redundant tokens at different VLM inference stages~\citep{aneja2026phi,beyer2024paligemma,team2025gemma}. FastV prunes visual tokens using attention scores from early language-model layers, whereas VisionZip preserves informative, complementary content before language-model processing~\citep{chen2024image,yang2025visionzip}. DivPrune, DART, and SparseVLM instead emphasize diversity-, duplication-, or sparsification-based selection~\citep{alvar2025divprune,wen2025stop,zhang2025sparsevlm}. These methods are compared on post-pruning accuracy and inference speed at several fixed retained-token budgets~\citep{hu2026illava,deng2026scope}. Yet a fixed budget does not define a comparable operating point: selectors may preserve different amounts of task-relevant information, while runtime depends on pruning stage, selection overhead, and computation incurred before pruning~\citep{li2025balanced,wang2026each,Chen_2026_CVPR,Sun_2026_CVPR,Yang_2025_ICCV}. Thus, equal token counts imply neither equal predictive performance nor equal computational cost, so fixed-budget comparisons may obscure the performance–efficiency trade-off across methods.

\textbf{Evaluation of visual-token pruning.}
Recent work has attempted to improve this evaluation paradigm~\citep{dong2026mmtok,baek2026agilepruner}. VTC-Bench studies compression sensitivity by focusing on examples solvable by the dense model, while EffiVLM-Bench incorporates behavioral agreement and measured inference speed~\citep{liao2026we,wang2025effivlm}. Nevertheless, aggregate scores can still mix correct-to-wrong losses with wrong-to-correct gains, and pruning-insensitive examples may dominate benchmark results. This leaves a gap for an evaluation protocol that directly connects token-selection quality, correctness preservation, and realized acceleration.
\section{A Closer Look at Token-Pruning Benchmarks:
Accuracy and Efficiency}
\label{sec:diagnosis}

We revisit visual token-pruning evaluation from its fundamental objective: identifying the method--configuration that maximizes realized inference speedup under a common performance requirement,
\begin{equation}
\max_{M\in\mathcal{M},\, c\in\mathcal{C}_M}
\underbrace{
    \frac{T_{\mathrm{orig}}}{T_p(M,c)}
}_{\operatorname{Speedup}(M,c)}
\qquad
\mathrm{s.t.}\qquad
\operatorname{Acc}_{\mathcal{D}}(M,c) \geq \tau,
\label{eq:pruning_speedup_objective}
\end{equation}
where $M\in\mathcal{M}$ indexes a pruning method and $c\in\mathcal{C}_M$ one of its evaluated configurations. $T_{\mathrm{orig}}$ and $T_p(M,c)$ are the measured runtimes of the original and pruned pipelines, with $T_p$ including pruning overhead; $\operatorname{Acc}_{\mathcal{D}}(M,c)$ is the post-pruning accuracy on $\mathcal{D}$, and $\tau$ the shared accuracy requirement. 

\textit{This objective requires evaluating two complementary dimensions: \textbf{accuracy} and \textbf{realized efficiency}.} The validity of this comparison therefore depends on (1) whether $\operatorname{Acc}_{\mathcal{D}}$ faithfully measures capability preservation, on examples informative for token-selection quality, and (2) whether $T_p$ reflects the realized cost of the complete pruning pipeline. We examine these requirements in turn.



\subsection{On Accuracy: Preservation and Selection Discriminability}
\label{sec:observation_preservation}

We first examine the accuracy side of Eq.~\ref{eq:pruning_speedup_objective}.
Existing token-pruning evaluations typically use aggregate post-pruning
accuracy as the performance criterion. To understand what this score actually
reflects, we decompose it according to how pruning changes the original
model's predictions:
\begin{equation}
\operatorname{Acc}_{\mathcal D}(M,c)
=
\underbrace{
\alpha_{\mathcal D}^{\mathrm{orig}}
}_{\text{\scriptsize original accuracy}}
-
\underbrace{
\alpha_{\mathcal D}^{\mathrm{orig}}
r_{C\rightarrow W}^{M,c}
}_{\text{\scriptsize correct$\rightarrow$wrong}}
+
\underbrace{
\left(1-\alpha_{\mathcal D}^{\mathrm{orig}}\right)
r_{W\rightarrow C}^{M,c}
}_{\text{\scriptsize wrong$\rightarrow$correct}} .
\label{eq:accuracy_decomposition}
\end{equation}
Here, $\alpha_{\mathcal D}^{\mathrm{orig}}$ is the original-model accuracy,
while $r_{C\rightarrow W}^{M,c}$ and $r_{W\rightarrow C}^{M,c}$ are the
rates of originally correct predictions becoming wrong and originally wrong
predictions becoming correct after pruning.

This decomposition explicitly illustrates how aggregate accuracy combines preservation losses and correction gains. Under this view, aggregate accuracy can fail in two distinct ways: corrections may inflate the score without reflecting capability preservation, while preserved predictions may still provide little evidence of effective token selection, as depicted below.

\paragraph{\textit{Challenge I: Wrong-to-correct gains can mask capability loss in aggregate accuracy.}}
To quantify the transition terms in Eq.~\ref{eq:accuracy_decomposition}, we measure correct-to-wrong and wrong-to-correct rates on LLaVA-1.5 and Qwen3-VL-8B at 20\% visual-token retention. As in Figure~\ref{fig:evaluation_diagnostics}(a), wrong-to-correct transitions are of a similar magnitude to correct-to-wrong transitions across methods. On LLaVA-1.5, the wrong-to-correct contribution is about half to two-thirds as large as the correct-to-wrong contribution, while the two are even closer on Qwen3-VL-8B (7.76\% vs.~8.85\%). These results show that
\textit{correction gains can mask over half---and sometimes nearly all---of the accuracy loss}.



These correction gains, however, need not reflect better token selection. We identify two mechanisms through which such gains can arise:
\begin{enumerate}[leftmargin=*]
    \item \textit{Accidental removal of visual distractors.} Pruning may incidentally remove visual evidence that mislead the original model. The stop-sign example in Figure~\ref{fig:evaluation_gaps}(a) illustrates this case:  pruning removes tokens that drive the original model toward an incorrect prediction, causing the pruned model to instead output the correct ``4-way'' answer. The correction therefore reflects incidental removal of misleading evidence rather than more effective visual evidence preservation.
    
    \item \textit{Increased reliance on textual priors.} As visual evidence is weakened, the model may rely more heavily on image-agnostic textual priors. We define stable corrections as examples that remain corrected across all random-pruning trials and compare them with a text-only control. As shown in Figure~\ref{fig:evaluation_diagnostics}(b), both the number of stable corrections and their overlap with text-only-correct examples increase as fewer visual tokens are retained. In particular, the overlap rises from about 45\% at 196 tokens to 76\% at 16 tokens. These corrections therefore reflect increased reliance on image-agnostic textual priors, rather than visually grounded evidence. 
    
\end{enumerate}

Together, these mechanisms show that wrong-to-correct gains can arise without improved token selection. However, aggregate accuracy allows spurious correction gains to offset losses on originally correct predictions, thereby underestimating the capability lost after pruning. 
Capability preservation should therefore be evaluated independently of pruning-induced correction gains.

\paragraph{\textit{Challenge~II: Pruning-insensitive examples obscure token-selection quality.}}
Even after separating capability preservation from correction gains, some examples remain correct regardless of which visual tokens are retained, making their preserved correctness weak evidence of effective token selection. We therefore ask a counterfactual: would the model still answer correctly if the retained tokens were chosen at random? We study this sensitivity via Random-Pass@$k$ (Eq.~\ref{eq:random_pass}), which counts how many of $k$ random-pruning trials preserve correctness for each originally correct example.


\begin{proposition}[\textit{High random-pruning success limits gains from informed selection}]
\label{prop:random_selection_difficulty}
Consider $M$ visual patches, $N$ of which independently suffice for a correct prediction.
If $B$ patches are retained uniformly at random and correctness requires retaining at least one such patch, then
\begin{equation}
p_N
=
1-\frac{\binom{M-N}{B}}{\binom{M}{B}},
\qquad
p_{N+1}\geq p_N,
\qquad
q_N-p_N\leq1-p_N,
\label{eq:random_evidence_redundancy}
\end{equation}
where $p_N$ and $q_N$ denote the success probabilities of random and informed selection.
Random pruning fails only when all $B$ retained patches come from the $M-N$ non-evidence patches, yielding $p_N$.
Increasing $N$ reduces this failure probability, while $q_N\leq1$ directly gives $q_N-p_N\leq1-p_N$.
\end{proposition}

Random-Pass@$k$ is therefore grounded by Proposition~\ref{prop:random_selection_difficulty}, which shows that higher random-pruning success leaves less room for informed selection to improve. With $k=10$, we accordingly group examples into Easy (8--10 random successes), Medium (4--7), and Hard (0--3).

As shown in Figure~\ref{fig:evaluation_gaps}(b), we reveal that Easy examples constitute over 80\% of MMBench-EN, POPE, and MME, causing pruning-insensitive samples to dilute differences among token selectors. Conversely, TextVQA contains 48.49\% Hard examples (Appendix~\ref{app:dataset-difficulty-distributions}), providing an explanation for its stronger discriminability observed in prior work. However, TextVQA covers only a narrow range of text-centric VLM capabilities, whereas \textit{\textbf{Random-Pass@$k$ enables broader benchmarks to be restructured into selection-sensitive evaluations}} while retaining the original capability coverage.




\subsection{On Realized Efficiency}
\label{sec:observation_acceleration}

\paragraph{\textit{Challenge III: Fixed-token comparisons misrepresent realized efficiency across methods.}}
Existing evaluations commonly compare methods at matched retained-token budgets. However, the same token budget does not place methods at comparable operating points. This creates two mismatches: \textit{(i) accuracy}, because the same token budget can correspond to different preservation levels across methods; and \textit{(ii) realized efficiency}, because identical token counts can incur different end-to-end costs depending on pruning location and selection overhead. Fixed-token comparisons can therefore misrepresent both predictive preservation and realized efficiency across methods.	

Figure~\ref{fig:evaluation_gaps}(c) confirms both mismatches. For preservation, CDPruner and FastV require only 16 and 270 visual tokens, respectively, to reach the same 95\% correctness-preservation target, showing that the same preservation can require very different token budgets. For realized efficiency, CDPruner retains fewer tokens than DivPrune (16 \textit{vs.} 52) but achieves a lower speedup ($1.98\times$ \textit{vs.} $2.08\times$), revealing that token compression and realized speedup can rank methods in opposite orders.

Cross-method comparison should therefore match preservation rather than token count and measure realized end-to-end speedup directly. This leads to comparing the maximum realized speedup under a shared preservation target, introduced as \emph{Preservation-Constrained Speedup} (PCS) in Section~\ref{sec:diagnosis_acceleration}.

\section{PAA-Bench: Evaluation Framework}
\label{sec:evaluation_framework}

The preceding analysis motivates an evaluation framework that separates token-selection quality from realized efficiency. Based on these observations, we introduce PAA-Bench with two complementary evaluation components: \textbf{Correctness Preservation and Selection-Difficulty Stratification} (Section~\ref{sec:preservation_selection}) and \textbf{Preservation-Constrained Speedup} (Section~\ref{sec:diagnosis_acceleration}).


\begin{algorithm}[t]
\caption{PAA-Bench evaluation protocol}
\label{alg:paa_bench}
\small
\begin{algorithmic}[1]
\Statex \textbf{Input:} Evaluation set $\mathcal D$, original VLM $f_D$,
pruning methods $\mathcal M$, random trials $k$, preservation target $\tau$
\Statex \textbf{Output:} Difficulty subsets, correctness preservation, and
$\{\PCS_M@\tau\}_{M\in\mathcal M}$

\Statex \textbf{Stage 1: Construct the preservation set}
\State $\mathcal D^+ \gets
\{x_i\in\mathcal D \mid f_D(x_i)=y_i\}$

\Statex \textbf{Stage 2: Stratify token-selection difficulty}
\For{each $x_i\in\mathcal D^+$}
    \State Compute $\operatorname{Random\text{-}Pass@}k(x_i)$ using $k$ random-pruning trials
    \State Assign $x_i$ to $\mathcal D_{\mathrm{Easy}}$,
    $\mathcal D_{\mathrm{Medium}}$, or $\mathcal D_{\mathrm{Hard}}$
\EndFor

\Statex \textbf{Stage 3: Compare realized efficiency at matched preservation}
\For{each pruning method $M\in\mathcal M$}
    \State Evaluate correctness preservation and realized speedup
    for each $c\in\mathcal C_M$
    \State $\displaystyle
    \PCS_M@\tau \gets
    \max_{c\in\mathcal C_M}
    \operatorname{Speedup}(M,c)
    \quad
    \mathrm{s.t.}\quad
    \operatorname{Acc}_{\mathcal D^+}(M,c)\geq\tau$
\EndFor
\end{algorithmic}
\end{algorithm}

\subsection{Correctness Preservation and Selection-Difficulty Stratification}
\label{sec:preservation_selection}



Section~\ref{sec:observation_preservation} shows that post-pruning correctness alone is insufficient to characterize token-selection quality: it may not represent preserved capability, and even preserved correctness may not be attributable to effective token selection, when the example itself is pruning-insensitive. We therefore evaluate token selection in two stages: first measuring correctness preservation on the original-correct subset, and then stratifying these examples by random-pruning sensitivity.


\textbf{Correctness preservation.}
As shown in Section~\ref{sec:diagnosis}, aggregate
post-pruning accuracy inherently conflates correct-to-wrong regressions with
wrong-to-correct gains. Since the latter may reflect distractor-removal
effects or textual-prior reliance, they can partially mask substantial losses
of originally correct predictions. This motivates explicitly conditioning
preservation on dense-model correctness, thereby excluding correction gains
from the resulting metric by construction.

Accordingly, let
$\Dplus=\{(x_i,y_i)\in\mathcal D\mid f_D(x_i)=y_i\}$
denote this dense-correct subset. For pruning method $M$ under configuration
$c$, we define correctness preservation as
\begin{equation}
\operatorname{Acc}_{\Dplus}(M,c)
=
\frac{
\sum_{(x_i,y_i)\in\Dplus}
\ind\!\left[f_{M,c}(x_i)=y_i\right]
}{
|\Dplus|
}.
\label{eq:correctness_preservation}
\end{equation}
This quantity measures the fraction of dense-correct examples that remain
correct after pruning. Higher values therefore indicate that the pruned model
preserves a larger share of the dense model's originally correct predictions,
while wrong-to-correct transitions are excluded by construction.

\textbf{Selection-difficulty stratification.}
After restricting evaluation to dense-correct examples, preservation alone
still does not establish effective token selection, since some examples remain
correct regardless of which visual tokens are retained. We therefore use
repeated random pruning to calibrate per-example selection sensitivity. For
each $(x_i,y_i)\in\Dplus$, we conduct $k$ independent trials under matched
pruning settings and define
\begin{equation}
\operatorname{Random\text{-}Pass@}k(x_i)
=
\sum_{j=1}^{k}
\ind\!\left[f_{m_j}(x_i)=y_i\right],
\qquad (x_i,y_i)\in\Dplus,
\label{eq:random_pass}
\end{equation}
where $m_j$ denotes the random pruning mask in trial $j$.
Random-Pass@$k$ ranges from $0$ to $k$; lower scores indicate that random
pruning is less likely to preserve correctness, creating more room for
performance differences among informed selectors to emerge, whereas higher
scores imply limited headroom and weaker discriminative value. With $k=10$,
we stratify examples into \textbf{Easy (8--10)}, \textbf{Medium (4--7)},
and \textbf{Hard (0--3)}, and report correctness preservation for each stratum. Together, correctness preservation quantifies retention
of the dense model's originally correct predictions, while Random-Pass@$k$
identifies selection-sensitive examples that better differentiate token
selectors.

\subsection{Preservation-Constrained Speedup}
\label{sec:diagnosis_acceleration}

Having established what pruning preserves and when token choice matters, we
next ask whether token reduction translates into realized efficiency. As shown
in Section~\ref{sec:observation_acceleration}, retained-token count alone cannot
characterize this trade-off, since methods differ in correctness preservation
and end-to-end cost. We therefore compare complete pruning pipelines under a
shared preservation target.

To this end, we introduce \emph{Preservation-Constrained Speedup} (PCS),
defined as the maximum measured end-to-end speedup over a method's evaluated
configurations satisfying a shared preservation target. For method $M$, we
define
\begin{equation}
\PCS_M@\tau
=
\max_{c\in\mathcal C_M}
\frac{T_D}{T_{M,c}}
\qquad
\mathrm{s.t.}\qquad
\operatorname{Acc}_{\Dplus}(M,c)\geq\tau,
\label{eq:pcs}
\end{equation}
where $\mathcal C_M$ denotes the set of evaluated configurations, and $T_D$
and $T_{M,c}$ denote the measured end-to-end runtimes of the dense and pruned
pipelines, respectively. Both are measured under matched hardware and inference
settings, with $T_{M,c}$ covering the complete pruning pipeline, including
token-selection and gathering overhead~\citep{wang2025effivlm}. PCS@95 sets
$\tau=0.95$ and identifies the fastest measured configuration that preserves
at least 95\% of the dense model's originally correct predictions, shifting
comparison from common token budgets to common preservation levels.
\section{Experiments}
\label{sec:experiments}
\suppressfloats[t]

Building on the preceding analysis of visual-token pruning evaluation, we conduct extensive experiments to systematically assess pruning methods across different evaluation criteria and quantify the efficiency gains realized in practical settings. We address the following research questions (\textbf{\textit{RQs}}):
\begin{list}{}{%
    \settowidth{\labelwidth}{\textbf{RQ3.}}%
    \setlength{\labelsep}{0.6em}%
    \setlength{\leftmargin}{\labelwidth}%
    \addtolength{\leftmargin}{\labelsep}%
    \setlength{\itemindent}{0pt}%
    \setlength{\listparindent}{0pt}%
    \setlength{\itemsep}{3pt}%
    \setlength{\parsep}{0pt}%
    \setlength{\topsep}{3pt}%
    \setlength{\partopsep}{0pt}%
    \renewcommand{\makelabel}[1]{\makebox[\labelwidth][l]{#1}}%
}
\item[\textit{\textbf{RQ1.}}] \emph{How do wrong-to-correct gains influence the relative rankings of token-pruning methods?}

\item[\textit{\textbf{RQ2.}}] \emph{Do performance differences among methods widen as token-selection difficulty increases?}

\item[\textit{\textbf{RQ3.}}] \emph{Which methods achieve the greatest realized speedup under a shared preservation target?}
\end{list}

\noindent\textbf{Setup.} Across three VLM backbones, LLaVA-1.5~\citep{liu2024improved}, LLaVA-NeXT~\citep{liu2024llavanext}, and Qwen3-VL~\citep{bai2025qwen3vl}, we evaluate eight pruning methods: CDPruner~\citep{zhang2026beyond}, DivPrune~\citep{alvar2025divprune}, FastV~\citep{chen2024image}, VisionZip~\citep{yang2025visionzip}, DART~\citep{wen2025stop}, PruMerge~\citep{shang2025prumerge}, SparseVLM~\citep{zhang2025sparsevlm}, and VisPruner~\citep{zhang2025vispruner}, with random pruning (Random) as a reference. Evaluation covers GQA~\citep{hudson2019gqa}, VizWiz~\citep{gurari2018vizwiz}, ScienceQA~\citep{lu2022scienceqa}, TextVQA~\citep{singh2019towards}, POPE~\citep{li2023evaluating}, MME~\citep{fu2025mme}, MMBench~\citep{liu2024mmbench}, and VQAv2~\citep{goyal2017making}, with method and task coverage varying by backbone. 
\begin{table}[htbp]
\caption{\textbf{Correctness preservation on LLaVA-1.5 (\%).} Dataset scores average the three difficulty strata. S-Avg and W-Avg average stratum-weighted and sample-weighted scores across nine datasets. Rel. averages pruned-to-dense score ratios over the same nine datasets. Bold/underline: best/second-best per budget. Full results appear in Appendix~\ref{app:selection_results}. }
\label{tab:llava_selection_main}
\centering
\vspace{4pt}
\definecolor{bandgray}{gray}{0.88}
\begingroup
\small
\newcommand{\paahead}[1]{\makebox[\linewidth][c]{\fontsize{8}{9}\selectfont\bfseries #1}}
\setlength{\tabcolsep}{2.8pt}\setlength{\arrayrulewidth}{0.3pt}
\renewcommand{\arraystretch}{1.04}
\newcommand{\paacellwidth}{\dimexpr(\linewidth-48pt-24\tabcolsep-2\arrayrulewidth)/12\relax}
\noindent\begin{tabular}{@{}>{\raggedright\arraybackslash}p{48pt}|*{9}{>{\centering\arraybackslash}p{\paacellwidth}}|*{3}{>{\centering\arraybackslash}p{\paacellwidth}}@{}}

\toprule[0.55pt]
{\fontsize{8}{9}\selectfont\bfseries Method} & \paahead{GQA} & \paahead{VizWiz} & \paahead{SQA\textsuperscript{I}} & \paahead{VQA\textsuperscript{T}} & \paahead{POPE} & \paahead{MME} & \paahead{VQA\textsuperscript{v2}} & \paahead{MMB} & \paahead{MMB\textsuperscript{CN}} & \paahead{S-Avg} & \paahead{W-Avg} & \paahead{Rel.} \\
\specialrule{0.3pt}{0.4pt}{0pt}
\rowcolor{bandgray}
\multicolumn{13}{c}{\textit{Retain 128 visual tokens}} \\
\specialrule{0pt}{0pt}{0pt}
Random & 56.7 & 56.3 & 55.2 & 55.1 & 57.8 & 56.5 & 57.3 & 55.3 & 54.3 & 56.1 & 85.6 & 90.2 \\
CDPruner & \underline{73.8} & 74.4 & 71.7 & 86.7 & 68.9 & 73.7 & 80.0 & 69.2 & 70.1 & 74.3 & 92.1 & \textbf{98.3} \\
DivPrune & 68.8 & 72.7 & 70.1 & 83.3 & 68.7 & 69.5 & 74.9 & 66.8 & 66.2 & 71.2 & 91.2 & \underline{97.4} \\
FastV & 63.3 & 81.1 & \textbf{85.2} & \underline{89.7} & 73.1 & \textbf{83.2} & 75.4 & 77.2 & 82.1 & 78.9 & 90.9 & 93.5 \\
DART & 72.8 & \underline{83.8} & \underline{85.0} & 86.1 & \underline{79.7} & \underline{81.3} & \underline{80.9} & \textbf{79.6} & \underline{83.3} & \underline{81.4} & \underline{93.0} & 96.6 \\
SparseVLM & \textbf{74.4} & \textbf{84.1} & 81.6 & \textbf{92.3} & \textbf{79.8} & 78.9 & \textbf{83.6} & \underline{78.2} & \textbf{85.6} & \textbf{82.1} & \textbf{94.4} & 96.4 \\
\specialrule{0.3pt}{0.4pt}{0pt}
\rowcolor{bandgray}
\multicolumn{13}{c}{\textit{Retain 64 visual tokens}} \\
\specialrule{0pt}{0pt}{0pt}
Random & 56.2 & 55.6 & 54.6 & 53.5 & 58.2 & 55.5 & 56.8 & 55.5 & 54.9 & 55.6 & 79.9 & 85.5\\
CDPruner & \underline{73.9} & 74.2 & 72.1 & \underline{84.3} & \underline{79.9} & 75.3 & \underline{82.5} & 75.1 & 72.0 & \underline{76.6} & \underline{89.3} & \textbf{96.9} \\
DivPrune & 69.2 & 72.0 & 70.7 & 82.4 & 79.1 & 69.4 & 76.5 & 66.0 & 68.4 & 72.6 & 88.0 & \underline{95.2} \\
FastV & 55.1 & 72.3 & \underline{81.9} & 82.2 & 58.7 & 71.5 & 65.6 & 72.2 & 74.6 & 70.5 & 83.1 & 76.5 \\
DART & 66.6 & \underline{77.4} & 80.3 & 79.6 & 71.5 & \underline{80.2} & 75.5 & \underline{75.4} & \underline{78.1} & 76.1 & 87.8 & 93.5 \\
SparseVLM & \textbf{76.5} & \textbf{84.1} & \textbf{86.1} & \textbf{92.0} & \textbf{82.7} & \textbf{83.0} & \textbf{86.8} & \textbf{86.1} & \textbf{90.1} & \textbf{85.3} & \textbf{93.3} & 88.1 \\
\specialrule{0.3pt}{0.4pt}{0pt}
\rowcolor{bandgray}
\multicolumn{13}{c}{\textit{Retain 32 visual tokens}} \\
\specialrule{0pt}{0pt}{0pt}
Random & 55.9 & 54.4 & 54.3 & 52.9 & 57.2 & 55.2 & 56.1 & 54.6 & 54.3 & 55.0 & 73.8 & 78.4 \\
CDPruner & \textbf{74.4} & \textbf{73.5} & \underline{74.2} & \textbf{80.7} & \textbf{82.9} & \textbf{74.0} & \textbf{82.9} & \textbf{75.0} & 67.4 & \textbf{76.1} & \textbf{85.5} & \textbf{94.9} \\
DivPrune & \underline{67.5} & \underline{70.5} & 71.8 & \underline{78.2} & \underline{78.0} & 70.6 & \underline{75.6} & 69.8 & \underline{71.5} & \underline{72.6} & \underline{83.9} & \underline{92.1} \\
DART & 61.7 & 69.0 & \textbf{77.4} & 71.7 & 68.8 & \underline{73.0} & 68.6 & \underline{70.7} & \textbf{72.1} & 70.3 & 80.7 & 89.6 \\
\bottomrule[0.55pt]
\end{tabular}
\endgroup

\vspace{-6pt}
\end{table}
\subsection{Method Rankings under Different Evaluation Metrics (RQ1)}
\label{sec:exp_rankings}
To test whether wrong-to-correct gains alter method rankings, 
Table~\ref{tab:llava_selection_main} compares RelAcc with W-Avg. 
RelAcc is the pruned-to-dense score ratio under each benchmark's original metric, 
whereas W-Avg measures correctness preservation on dense-correct examples, 
weighting difficulty strata by their sample counts within each dataset and 
averaging datasets equally. On LLaVA-1.5 with 64 retained visual tokens, 
CDPruner achieves a RelAcc of 96.9 versus SparseVLM's 88.1, but their W-Avg 
scores are 89.3 and 93.3, respectively, reversing the ranking. Thus, higher 
relative accuracy does not necessarily imply better preservation of the dense 
model's correct predictions.

The disagreement extends beyond the preceding example. Among the five
methods reported under both metrics, five of ten pairwise orderings
at 128 tokens and three at 64 tokens reverse between RelAcc and W-Avg. At both
budgets, SparseVLM moves from fourth under RelAcc to first under W-Avg. These
reversals highlight the diagnostic value of PAA-Bench: its preservation-aware
evaluation reveals substantial ranking differences obscured by conventional
aggregate task scores.

These ranking shifts reflect a fundamental distinction between the two
evaluation metrics. RelAcc allows wrong-to-correct gains to offset
correct-to-wrong losses in the benchmark score, whereas W-Avg conditions on
dense-correct examples and directly quantifies correctness preservation.
RelAcc therefore summarizes net task performance, while W-Avg explicitly
reveals pruning-induced losses of originally correct predictions; the former
should not be treated as a reliable proxy for the latter.

\subsection{Discriminability across Selection Difficulty (RQ2)}
\label{sec:exp_attribution}

\begin{figure}[!t]
    \centering
    \includegraphics[width=\linewidth]{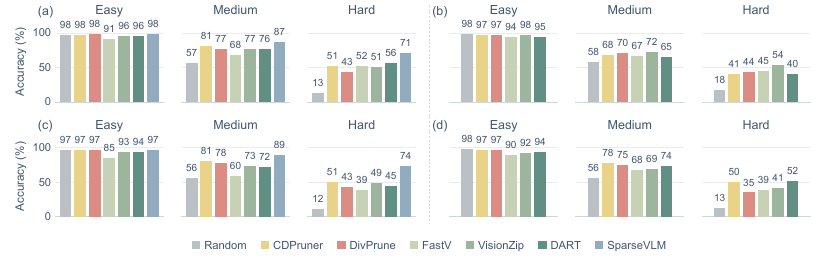}
    \caption{\textbf{Method differences widen with selection difficulty.}
    (a) LLaVA-1.5, 64 tokens; (b) Qwen3-VL, 30\% retention;
    (c) LLaVA-1.5, 32 tokens; (d) LLaVA-NeXT, 160 tokens.
    Random reports the mean of 10 trials.
    Implementation details are in Appendix~\ref{app:stratified_figure_settings}.}
    \vspace{-20pt}
    \label{fig:difficulty_stratified_accuracy}
\end{figure}
To assess how selection difficulty affects method separation,
Figure~\ref{fig:difficulty_stratified_accuracy} compares correctness
preservation across Easy, Medium, and Hard examples. Across all four
evaluated settings, the gap in correctness preservation among methods widens
with difficulty. We define separation as the difference between the highest
and lowest scores among the displayed pruning methods, excluding Random. On
LLaVA-1.5 with 64 retained visual tokens, this gap increases from 6.9
percentage points on Easy examples to 18.8 on Medium and 27.6 on Hard
examples. Taken together, these results show that harder examples provide a
more discriminative evaluation of token-selection strategies.

Tables~\ref{tab:llava_selection_main} and~\ref{tab:next_selection_main}
extend the difficulty-aware comparison across vision-language tasks using
W-Avg and S-Avg. Both weight datasets equally, but within each dataset
W-Avg preserves empirical stratum frequencies, whereas S-Avg weights Easy,
Medium, and Hard equally. For LLaVA-NeXT at 320 retained visual tokens, the
gap between Random and DART increases from 7.5 percentage points under W-Avg
(84.8 vs.\ 92.3) to 25.5 under S-Avg (56.0 vs.\ 81.5); the gap between
VisionZip and Random similarly grows from 4.2 to 14.4 points. When Easy
examples dominate the sample mix, empirical weighting can therefore obscure
method separation in the selection-sensitive strata. By reporting both
aggregates, PAA-Bench makes these difficulty-dependent differences more visible.

\subsection{Realized Speedup under Preservation Constraints (RQ3)}
\label{sec:exp_acceleration}
\begin{table}[!t]
\caption{\textbf{Correctness preservation on LLaVA-NeXT (\%).} Selected methods at 640 and 320 retained visual tokens. Dataset scores average the three difficulty strata; S-Avg and W-Avg average stratum-weighted and sample-weighted scores across eight datasets. Bold/underline: best/second-best among the shown methods per budget. Full results appear in Appendix~\ref{app:selection_results}.}
\label{tab:next_selection_main}
\centering
\vspace{4pt}
\definecolor{bandgray}{gray}{0.88}
\begingroup
\small
\newcommand{\paahead}[1]{\makebox[\linewidth][c]{\fontsize{8}{9}\selectfont\bfseries #1}}
\setlength{\tabcolsep}{2.8pt}\setlength{\arrayrulewidth}{0.3pt}
\renewcommand{\arraystretch}{1.04}
\newcommand{\paacellwidth}{\dimexpr(\linewidth-48pt-22\tabcolsep-2\arrayrulewidth)/11\relax}
\noindent\begin{tabular}{@{}>{\raggedright\arraybackslash}p{48pt}|*{8}{>{\centering\arraybackslash}p{\paacellwidth}}|*{3}{>{\centering\arraybackslash}p{\paacellwidth}}@{}}
\toprule[0.55pt]
{\fontsize{8}{9}\selectfont\bfseries Method} & \paahead{GQA} & \paahead{VizWiz} & \paahead{SQA\textsuperscript{I}} & \paahead{VQA\textsuperscript{T}} & \paahead{POPE} & \paahead{MME} & \paahead{MMB} & \paahead{MMB\textsuperscript{CN}} & \paahead{S-Avg} & \paahead{W-Avg} & \paahead{Rel.} \\

\specialrule{0.3pt}{0.4pt}{0pt}
\rowcolor{bandgray}
\multicolumn{12}{c}{\textit{Retain 640 visual tokens}} \\
\specialrule{0pt}{0pt}{0pt}
Random & 57.2 & 56.0 & 56.9 & 55.3 & 59.1 & 58.0 & 56.3 & 56.3 & 56.9 & 89.6 & 93.6 \\
DART & \textbf{83.2} & \textbf{86.8} & \textbf{86.0} & \textbf{90.3} & \textbf{89.1} & \textbf{86.3} & \textbf{80.7} & \textbf{87.6} & \textbf{86.3} & \textbf{96.1} & \underline{99.1} \\
DivPrune & 62.5 & 69.8 & \underline{68.7} & 75.1 & \underline{75.0} & \underline{75.1} & 67.9 & 58.9 & 69.1 & 91.7 & 99.0 \\
VisionZip & \underline{64.7} & \underline{75.1} & 63.8 & \underline{86.0} & 73.5 & 72.5 & \underline{73.2} & \underline{66.3} & \underline{71.9} & \underline{92.7} & \textbf{100.1} \\
\specialrule{0.3pt}{0.4pt}{0pt}
\rowcolor{bandgray}
\multicolumn{12}{c}{\textit{Retain 320 visual tokens}} \\
\specialrule{0pt}{0pt}{0pt}
Random & 56.6 & 55.0 & 56.2 & 53.7 & 58.3 & 56.2 & 57.1 & 55.2 & 56.0 & 84.8 & 89.7 \\
DART & \textbf{77.4} & \textbf{82.1} & \textbf{77.6} & \textbf{85.4} & \textbf{81.7} & \textbf{82.6} & \textbf{79.1} & \textbf{85.9} & \textbf{81.5} & \textbf{92.3} & \underline{97.0} \\
DivPrune & \underline{63.8} & 69.4 & \underline{62.1} & 74.5 & \underline{80.4} & \underline{70.6} & \underline{73.7} & \underline{64.6} & 69.9 & 88.9 & \textbf{97.3} \\
VisionZip & 63.5 & \underline{75.6} & 61.2 & \underline{85.0} & 75.1 & 68.4 & 70.4 & 63.9 & \underline{70.4} & \underline{89.0} & 96.8 \\
\bottomrule[0.55pt]
\end{tabular}
\endgroup

\vspace{-6pt}
\end{table}
\begin{figure}[t]
\centering
\includegraphics[width=\linewidth]{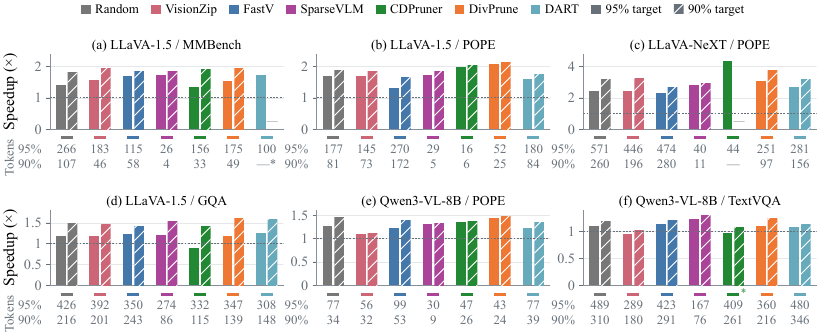}
\caption{\textbf{Fewer tokens do not guarantee faster inference.}
Bars show end-to-end speedup at nominal 95\% and 90\% dense-correct preservation targets; numbers below report retained visual tokens. Dashed lines mark the dense baseline ($1\times$). Y-axis ranges vary.}
\label{fig:pcs_representative}
\vspace{-15pt}
\end{figure}
We use PCS to compare measured end-to-end speedup under shared
correctness-preservation targets.
Figure~\ref{fig:pcs_representative} compares measured results across methods, with complete results provided in Appendix~\ref{app:pcs_complete}. The leading method varies across models and tasks. At 95\% preservation, CDPruner achieves the highest speedup of $4.37\times$ on LLaVA-NeXT/POPE, whereas SparseVLM leads on Qwen3-VL/TextVQA with $1.23\times$ and reaches $1.31\times$ at 90\% preservation.

Greater token compression does not necessarily yield greater realized speedup. On LLaVA-1.5/MMBench at 90\% preservation, SparseVLM retains only 4 visual tokens and achieves a $1.83\times$ speedup, whereas DivPrune retains 49 visual tokens yet achieves a higher speedup of $1.95\times$. Pruning may even fail to produce a net acceleration: on Qwen3-VL/TextVQA at 95\% preservation, VisionZip retains 289 visual tokens but achieves a speedup of only $0.95\times$. As discussed in Section~\ref{sec:observation_acceleration}, runtime also depends on pruning location and selection overhead, making retained visual-token count insufficient to characterize the inference cost of the complete pruning procedure.

The correctness-preservation requirement can also change the speed ranking of methods. On LLaVA-1.5/MMBench, SparseVLM is faster than DivPrune at 95\% preservation, with speedups of $1.74\times$ and $1.53\times$, respectively. When the requirement is relaxed to 90\%, DivPrune reaches $1.95\times$ and surpasses SparseVLM's $1.83\times$. Comparisons of acceleration must therefore specify the correctness-preservation threshold. PCS reports the maximum measured speedup each method achieves among the evaluated configurations that satisfy a common threshold, thereby accounting jointly for token selection, pruning location, and algorithmic overhead.
\section{Conclusion}
\label{sec:conclusion}

Visual-token pruning should be evaluated by what it preserves, when token
choice matters, and how much end-to-end acceleration it delivers in practical
settings. PAA-Bench addresses these goals through dense-correct preservation,
Random-Pass@$k$-based selection-difficulty stratification, and PCS under shared
preservation targets. Across eight methods, three VLM backbones, and nine
benchmarks, preservation-aware scoring changes method rankings,
selection-sensitive examples expose clearer differences, and stronger token
compression does not necessarily yield greater speedup. Efficiency rankings
vary across tasks and preservation targets. Together, these findings support
comparing pruning pipelines at matched preservation levels and reporting
results across selection-difficulty strata, rather than relying on aggregate
scores or retained-token counts alone.

\clearpage
\subsection*{AI Use Statement}
Generative AI was used to support the organization and development of the
manuscript from a provided research outline, including drafting initial prose,
restructuring sections, refining sentence flow, and improving clarity,
coherence, and terminological consistency. It also assisted with preliminary
reference discovery by identifying potentially relevant literature for further
review. 

\subsection*{Reproducibility Statement}

The definitions of the proposed metrics and the complete evaluation
protocol are provided in Section 4 and Appendix A. Dataset and benchmark
coverage are described in Appendix B, while model configurations, token budgets, random interventions, and runtime measurement
boundaries are documented in Appendix C. Complete token-selection and
speedup results are reported in Appendices D and E, with qualitative
records and sensitivity analyses provided in Appendices F and G.


\clearpage
\appendix
\begin{center}
{\Large\bfseries Supplementary Material}
\end{center}
\noindent This appendix fills in the evaluation details that are not developed in the main paper. Sections~\ref{app:framework_details}--\ref{app:protocol} clarify the metric definitions, benchmark coverage, and experimental protocol; Sections~\ref{app:selection_results} and~\ref{app:pcs_complete} report the complete token-selection and PCS results; Section~\ref{app:qualitative} presents qualitative cases with their recorded random-pruning outcomes; and Section~G examines how Random-Pass strata change across token budgets and how sensitive the assignments are to the number of random trials.

\bigskip
\noindent{\large\bfseries Contents}\par\medskip
\makeatletter
\begingroup
\hypersetup{linkcolor=black}
\small
\setlength{\parskip}{0pt}
\setcounter{tocdepth}{2}
\renewcommand{\l@section}[2]{%
  \addvspace{6pt}%
  \begingroup\bfseries\@dottedtocline{1}{0em}{2.1em}{#1}{#2}\endgroup}
\renewcommand{\l@subsection}[2]{\@dottedtocline{2}{1.5em}{2.7em}{#1}{#2}}
\@starttoc{apc}
\endgroup
\makeatother
\clearpage

\let\paaAppendixSection\section
\renewcommand{\section}[1]{%
  \paaAppendixSection{#1}%
  \addcontentsline{apc}{section}{\protect\numberline{\thesection}#1}}
\let\paaAppendixSubsection\subsection
\renewcommand{\subsection}[1]{%
  \paaAppendixSubsection{#1}%
  \addcontentsline{apc}{subsection}{\protect\numberline{\thesubsection}#1}}

\section{Additional Details of the Evaluation Framework}
\label{app:framework_details}

\subsection{Correctness Preservation and Prediction Transitions}
Aggregate accuracy records only the net change after pruning. Some predictions that the dense model gets right may become wrong, while some predictions that it gets wrong may become correct, and the two effects can offset each other. For a fixed model and dataset, let $b_i^D,b_i^P\in\{0,1\}$ denote the correctness of the dense and pruned predictions under the same answer evaluator. Define $\Dplus=\{i:b_i^D=1\}$ and $\Dminus=\{i:b_i^D=0\}$. Each example then falls into one of four transitions: correct-to-correct, correct-to-wrong, wrong-to-correct, or wrong-to-wrong. Over $N$ examples,
\begin{equation}
\operatorname{Acc}_P-\operatorname{Acc}_D
=\frac{N_{W\rightarrow C}-N_{C\rightarrow W}}{N}.
\label{eq:app_accuracy_transition}
\end{equation}
A small net accuracy drop can therefore hide many lost correct predictions when wrong-to-correct transitions offset them. To measure preservation directly, we condition on the dense model being correct:
\begin{equation}
\operatorname{Acc}_{\Dplus}(P)
=\frac{N_{C\rightarrow C}}{|\Dplus|}
=1-\frac{N_{C\rightarrow W}}{|\Dplus|},
\qquad |\Dplus|>0.
\label{eq:app_preservation}
\end{equation}
PCS constrains this quantity. A preservation rate of 95\% means that the pruned model keeps 95\% of the predictions that the dense model originally answered correctly. It does not mean 95\% accuracy on the full dataset, so preservation should be read alongside the dense model's task score.

\paragraph{Relation to relative accuracy.}
When both scores are binary accuracies on the same examples and $\operatorname{Acc}_D>0$,
\begin{equation}
\frac{\operatorname{Acc}_P}{\operatorname{Acc}_D}
=\operatorname{Acc}_{\Dplus}(P)
+\frac{N_{W\rightarrow C}}{|\Dplus|}.
\label{eq:app_relacc}
\end{equation}
The second term is the contribution from wrong-to-correct transitions. Relative accuracy can therefore rise because pruning corrects dense-model errors, even when it does not preserve more of the dense model's correct predictions.

\subsection{Random-Pass Stratification and Its Interpretation}
Correctness preservation tells us how much survives pruning, but not whether the retained tokens were chosen well. To measure how much token choice matters for each example, we hold the model, token budget, pruning location, eligible positions, and decoding rule fixed, and vary only the random mask. Let $p_i$ be the probability that a random mask preserves correctness on dense-correct example $i$. Under independent and identically distributed trials,
\begin{equation}
R_i=\operatorname{Random\text{-}Pass@}k(x_i)\sim\operatorname{Binomial}(k,p_i),
\quad \mathbb E[R_i/k]=p_i,
\quad \operatorname{Var}(R_i/k)=\frac{p_i(1-p_i)}{k}.
\label{eq:app_random_binomial}
\end{equation}
Random-Pass@$k$ records the \emph{number of successful trials} among $k$ random masks; it is not the probability of succeeding at least once. We use $k=10$ and define Easy as 8--10 successes, Medium as 4--7, and Hard as 0--3. These labels describe sensitivity to random token selection under the specified intervention, not an intrinsic difficulty level of the original benchmark question.

\paragraph{Why stratification can expose selection gains.}
Suppose an informed selector succeeds on example $i$ with probability $q_i\leq1$. Its gain over random selection satisfies $q_i-p_i\leq1-p_i$. When $p_i$ is already high, as it is for Easy examples, even a perfect selector has little room to improve over Random. Lower random success leaves more headroom. The wider method gaps in Figure~\ref{fig:difficulty_stratified_accuracy} are an empirical result of the evaluated settings, not something implied by the definition alone. 

\Needspace{10\baselineskip}
\subsection{Dataset and Stratum Aggregation}
\label{app:aggregation}
S-Avg and W-Avg answer different questions. S-Avg asks how a method performs when Easy, Medium, and Hard examples count equally, whereas W-Avg keeps the observed mixture of those strata within each dataset. For dataset $d$ and stratum $s\in\{E,M,H\}$, let $n_{d,s}$ be the number of dense-correct examples and $a_{d,s}$ the fraction preserved by a method. We first compute
\begin{equation}
S_d=\frac{a_{d,E}+a_{d,M}+a_{d,H}}{3},
\qquad
W_d=\frac{\sum_s n_{d,s}a_{d,s}}{\sum_s n_{d,s}}.
\label{eq:app_stratum_averages}
\end{equation}
For $J$ evaluated datasets, the reported percentage scores are
\begin{equation}
\mathrm{S\text{-}Avg}=\frac{100}{J}\sum_{d=1}^{J}S_d,
\qquad
\mathrm{W\text{-}Avg}=\frac{100}{J}\sum_{d=1}^{J}W_d.
\label{eq:app_dataset_averages}
\end{equation}
Both metrics give every dataset the same weight. W-Avg weights examples within each dataset, so it is not an example-pooled average over all datasets. S-Avg gives the three strata equal weight, so it is not accuracy under the benchmark's original sample distribution. The individual dataset columns in the complete token-selection tables report $100S_d$.

\subsection{From Measured Configurations to PCS}
The fixed-budget tables ask how much correctness a method preserves at a particular token budget. PCS asks a different question: among the configurations that meet a shared preservation target, which measured configuration is fastest?

We first obtain dense predictions and correctness labels, use random pruning to form the difficulty strata, and evaluate each candidate pruning configuration on the same examples. PCS then selects the fastest measured configuration whose correctness preservation reaches the target. This comparison measures the inference cost of achieving the target instead of treating the final retained-token count as the cost itself.

For a fixed candidate set, lowering the preservation target can only enlarge the feasible set. Whenever at least one feasible configuration exists,
\begin{equation}
\tau_1\leq\tau_2
\quad\Longrightarrow\quad
\PCS_M@\tau_1\geq\PCS_M@\tau_2.
\label{eq:app_pcs_monotonicity}
\end{equation}

\clearpage
\section{Datasets and Benchmark Coverage}
\label{app:datasets}

\subsection{Coverage and Task Roles}
The main token-selection tables cover nine benchmark variants for LLaVA-1.5-7B and eight each for LLaVA-NeXT-7B and Qwen3-VL-8B. The additional Qwen3-VL-2B and LLaVA-1.5-13B results cover seven variants and omit GQA and VQAv2. MMBench-EN and MMBench-CN are treated as separate variants, while VQAv2 appears only in the LLaVA-1.5-7B table. The benchmark set spans natural-image reasoning, text reading, science questions, object hallucination, broad perception, and questions collected in an assistive-use setting. Table~\ref{tab:app_datasets} summarizes the exact coverage used in the complete token-selection results. 

\begin{table}[htbp]
\centering
\caption{Benchmark variants in the complete token-selection tables. ``Primary three'' denotes LLaVA-1.5-7B, LLaVA-NeXT-7B, and Qwen3-VL-8B; ``All five'' additionally includes Qwen3-VL-2B and LLaVA-1.5-13B. Coverage here refers to the reported token-selection results, not every PCS or diagnostic experiment.}
\label{tab:app_datasets}
\small
\setlength{\tabcolsep}{4pt}
\renewcommand{\arraystretch}{1.12}
\begin{tabular}{@{}L{0.23\linewidth}L{0.46\linewidth}L{0.24\linewidth}@{}}
\toprule
Benchmark / table label & Main capability examined & Model coverage \\
\midrule
GQA & Objects, attributes, relations, and compositional reasoning & Primary three \\
VizWiz & Real-world visual questions from blind users & All five \\
ScienceQA-IMG / SQA$^I$ & Science questions with image context & All five \\
TextVQA / VQA$^T$ & Reading and reasoning about text in images & All five \\
POPE & Object existence and object hallucination & All five \\
MME & Perceptual and cognitive judgments & All five \\
VQAv2 / VQA$^{v2}$ & Open-ended natural-image question answering & LLaVA-1.5-7B \\
MMBench-EN / MMB & General multimodal perception and reasoning & All five \\
MMBench-CN / MMB$^{CN}$ & Chinese-language multimodal evaluation & All five \\
\bottomrule
\end{tabular}
\end{table}

\subsection{Natural-Image and Assistive Question Answering}
\paragraph{GQA.}
GQA~\citep{hudson2019gqa} builds compositional questions from scene graphs, including attributes, object relations, and multi-step reasoning. These questions are useful for token pruning because the answer may depend on several objects or on a relation between them; preserving only the most salient object may not be enough. The GQA column in our tables reports stratified correctness preservation, not the benchmark's original full-set accuracy.

\paragraph{VQAv2.}
VQAv2~\citep{goyal2017making} pairs open-ended questions with natural images and multiple human answers. Its complementary image pairs reduce the value of answering from the question alone, adding a broad natural-image task to the LLaVA-1.5-7B comparison. The native evaluator uses answer normalization and consensus-based fractional credit, whereas the transition analysis requires a binary correct/incorrect event. We keep those two quantities separate.

\paragraph{VizWiz.}
VizWiz~\citep{gurari2018vizwiz} contains user-captured images and spoken questions from blind users, together with multiple human answers and an explicit unanswerable category. It tests whether pruning preserves useful evidence in imperfect, real-world images without turning insufficient evidence into an overconfident answer. Preservation scores on this dataset therefore depend on how answer normalization and unanswerable cases are handled; the task is not simply central-object recognition.

\subsection{Text Reading and Science Questions}
\paragraph{TextVQA.}
TextVQA~\citep{singh2019towards} requires the model to read and reason about text in an image. A selector can preserve the broad scene and still remove a small sign or label that determines the answer, making this task particularly sensitive to fine visual evidence. Whether OCR text is supplied in the prompt also matters, because external text can reduce the model's dependence on the retained visual tokens.

\paragraph{ScienceQA-IMG.}
ScienceQA~\citep{lu2022scienceqa} contains multiple-choice science questions with image and/or text context; our tables use the image subset, denoted SQA$^I$. The task combines visual interpretation with scientific knowledge, but the presence of an image does not mean that every question requires it. Random-Pass strata add that missing information by measuring how strongly the evaluated model's answer depends on visual-token selection.

\subsection{Hallucination and Broad Multimodal Evaluation}
\paragraph{POPE.}
POPE~\citep{li2023evaluating} asks whether particular objects are present in an image and is designed to probe object hallucination. Its random, popular, and adversarial variants differ in how absent objects are sampled. For token pruning, POPE tests whether the evidence needed for a yes/no object judgment remains available after selection. A high POPE preservation score should not be read as evidence that fine-grained reading or relational reasoning is also preserved; those abilities are covered by other tasks.

\paragraph{MME.}
MME~\citep{fu2025mme} evaluates perception and cognition across 14 subtasks using short instruction--answer pairs. Its breadth complements the more focused benchmarks. The percentage in our token-selection tables is the stratum-balanced preservation score defined in Section~\ref{app:aggregation}, not the original MME composite score. Reproducing that value requires the same subtask set and the same unit for assigning binary correctness labels.

\paragraph{MMBench-EN and MMBench-CN.}
MMBench~\citep{liu2024mmbench} evaluates multimodal perception and reasoning with multiple-choice questions. We report the English and Chinese variants separately to show how methods behave across the two language settings. Differences between the columns cannot be attributed to visual selection alone, however; language understanding, answer extraction, and evaluator version can also affect them.

\clearpage
\section{Implementation and Evaluation Protocols}
\label{app:protocol}

\subsection{Models, Methods, and Budget Coverage}
The primary experiments use three backbones: LLaVA-1.5-7B, LLaVA-NeXT-7B, and Qwen3-VL-8B. The supplementary token-selection results add Qwen3-VL-2B and LLaVA-1.5-13B, without runtime measurements for either configuration. As in the main paper, all runs use a PRO 6000 GPU and batch size 1; methods within a comparison share the same prompts and a maximum output length of 128 tokens. 

The complete LLaVA-1.5-7B table includes CDPruner, DivPrune, FastV, VisionZip, DART, PruMerge, SparseVLM, and VisPruner~\citep{zhang2025vispruner}, plus Random. LLaVA-NeXT-7B includes the same set except VisPruner. The Qwen3-VL-8B selection table includes CDPruner, DivPrune, FastV, VisionZip, and DART, plus Random. Both additional model tables also report these five methods and Random.

\begin{table}[htbp]
\centering
\caption{Budgets and coverage of the complete fixed-budget selection tables. Qwen3-VL uses retention fractions rather than a common absolute token count.}
\label{tab:app_budget_coverage}
\small
\begin{tabular}{@{}lccc@{}}
\toprule
Model & Budgets & Benchmark variants & Methods + Random \\
\midrule
LLaVA-1.5-7B & 128, 64, 32 tokens & 9 & $8+1$ \\
LLaVA-NeXT-7B & 640, 320, 160 tokens & 8 & $7+1$ \\
Qwen3-VL-8B & 10\%, 20\% retention & 8 & $5+1$ \\
Qwen3-VL-2B & 20\%, 10\% retention & 7 & $5+1$ \\
LLaVA-1.5-13B & 128, 64, 32 tokens & 7 & $5+1$ \\
\bottomrule
\end{tabular}
\end{table}

\subsection{Dense References and Random Interventions}
Correctness preservation is a paired comparison. Each pruned prediction must be matched to the dense prediction for the same example, using the same prompt, image processing, answer evaluator, and decoding rule. Changing any of these choices can change $\Dplus$ and therefore the denominator of the preservation score. The random interventions use the same evaluation setup; only the token mask changes across trials.

\subsection{Settings of the Main-Paper Diagnostics}
\label{app:figures}
\paragraph{Difficulty-stratified comparison.}
\label{app:stratified_figure_settings}
Figure~\ref{fig:difficulty_stratified_accuracy} forms the Easy, Medium, and Hard strata separately for each model and token budget. For Qwen3-VL, the figure uses adapted implementations of CDPruner, FastV, VisionZip, and DART; SparseVLM appears only in the LLaVA-1.5 panels. Bar labels are rounded to integer percentages, while the bar heights use the unrounded values.

\paragraph{Transitions and correction overlap.}
\label{app:diagnostic_settings}
Figure~\ref{fig:evaluation_diagnostics}(a) uses 64 retained tokens for LLaVA-1.5 and 20\% retention for Qwen3-VL-8B. Panel~(b) averages the LLaVA-1.5 results equally across GQA, POPE, and MMBench-EN. A stable correction is a dense-incorrect example that becomes correct in all five random-pruning trials; an occasional correction succeeds in one to four trials. These five-trial correction groups are separate from the ten-trial difficulty strata defined on $\Dplus$. Panel~(c) uses 128 tokens for LLaVA-1.5, 320 for LLaVA-NeXT, and 20\% retention for Qwen3-VL-8B.

For a dense-incorrect example, we estimate the correction frequency as
\begin{equation}
\widehat p_i^{\,\mathrm{corr}}=\frac{1}{5}\sum_{j=1}^{5}b_i^{(j)},
\qquad i\in\Dminus.
\end{equation}
Text-only overlap is the fraction of examples in a correction group that are also answered correctly without image input. A higher overlap is consistent with greater reliance on textual knowledge, but it does not identify the cause of every correction. The control also removes the image before multimodal fusion, which is not the same intervention as deleting visual tokens after image information has already entered other representations.

\subsection{Runtime Boundaries and Configuration Selection}
Retained-token count captures only part of the cost of a pruning method. End-to-end timing must cover the same boundary for the dense and pruned systems, including the vision encoder, projector, language-model execution, and selection work such as scoring and token gathering. CPU preprocessing and device transfer should either be included for both systems or excluded for both and reported separately.

This matched boundary is especially important when pruning happens after several decoder layers: a method may retain few tokens at the end while still paying for substantial computation before pruning. PCS therefore uses measured pipeline time, including selection overhead, rather than inferring efficiency from the final token count.

\clearpage
\section{Complete Token-Selection Results and Analysis}
\label{app:selection_results}
Tables~\ref{tab:llava_retention_results}--\ref{tab:llava15_13b_selection} show the fixed-budget token-selection results, aggregated as described in Section~\ref{app:aggregation}. All scores are percentages, and all differences are reported in percentage points.

\subsection{LLaVA-1.5-7B}
\begin{table}[htbp]
\caption{LLaVA-1.5 results on nine benchmarks with 128, 64, and 32 retained visual tokens. S-Avg and W-Avg are equal-weight averages across datasets of the three-stratum and sample-weighted scores, respectively. Bold and underlined values indicate the best and second-best results, respectively, within each column and token budget.}
\label{tab:llava_retention_results}
\begin{center}
\definecolor{bandgray}{gray}{0.88}
\begingroup
\small
\newcommand{\paahead}[1]{{\small\bfseries #1}}
\setlength{\tabcolsep}{2.8pt}\setlength{\arrayrulewidth}{0.3pt}
\renewcommand{\arraystretch}{1.0}
\noindent\begin{tabular}{@{}l|ccccccccc|cc@{}}

\toprule[0.55pt]
\makebox[44pt][l]{\paahead{Method}} & \paahead{GQA} & \paahead{VizWiz} & \paahead{SQA\textsuperscript{I}} & \paahead{VQA\textsuperscript{T}} & \paahead{POPE} & \paahead{MME} & \paahead{VQA\textsuperscript{v2}} & \paahead{MMB} & \paahead{MMB\textsuperscript{CN}} & \paahead{S-Avg} & \paahead{W-Avg} \\
\specialrule{0.3pt}{0.4pt}{0pt}
\rowcolor{bandgray}
\multicolumn{12}{c}{\textit{Retain 128 visual tokens}} \\
\specialrule{0pt}{0pt}{0pt}
Random & 56.71 & 56.33 & 55.22 & 55.11 & 57.79 & 56.48 & 57.34 & 55.28 & 54.34 & 56.07 & 85.61 \\
CDPruner & \underline{73.75} & 74.38 & 71.68 & 86.74 & 68.89 & 73.69 & 80.01 & 69.16 & 70.09 & 74.27 & 92.10 \\
DivPrune & 68.80 & 72.74 & 70.11 & 83.31 & 68.68 & 69.48 & 74.86 & 66.80 & 66.18 & 71.22 & 91.21 \\
FastV & 63.29 & 81.11 & \textbf{85.23} & \underline{89.65} & 73.10 & \textbf{83.15} & 75.44 & 77.17 & 82.13 & 78.92 & 90.94 \\
VisionZip & 65.27 & 78.96 & 76.82 & 88.60 & 68.41 & 73.85 & 76.68 & 65.70 & 76.21 & 74.50 & 91.99 \\
DART & 72.75 & \underline{83.79} & \underline{85.03} & 86.08 & \underline{79.69} & \underline{81.34} & \underline{80.88} & \textbf{79.64} & \underline{83.25} & \underline{81.38} & \underline{92.96} \\
PruMerge & 61.82 & 73.25 & 71.02 & 80.77 & 69.89 & 73.38 & 74.28 & 65.94 & 72.03 & 71.38 & 90.10 \\
SparseVLM & \textbf{74.39} & \textbf{84.07} & 81.62 & \textbf{92.26} & \textbf{79.76} & 78.92 & \textbf{83.64} & \underline{78.18} & \textbf{85.63} & \textbf{82.05} & \textbf{94.43} \\
VisPruner & 65.98 & 78.85 & 75.60 & 89.61 & 71.89 & 74.98 & 76.87 & 70.29 & 71.34 & 75.05 & 92.25 \\
\specialrule{0.3pt}{0.4pt}{0pt}
\rowcolor{bandgray}
\multicolumn{12}{c}{\textit{Retain 64 visual tokens}} \\
\specialrule{0pt}{0pt}{0pt}
Random & 56.15 & 55.56 & 54.61 & 53.48 & 58.24 & 55.50 & 56.81 & 55.52 & 54.85 & 55.64 & 79.94 \\
CDPruner & \underline{73.93} & 74.15 & 72.12 & 84.32 & \underline{79.94} & 75.34 & \underline{82.52} & 75.11 & 71.98 & \underline{76.60} & \underline{89.33} \\
DivPrune & 69.16 & 72.01 & 70.69 & 82.36 & 79.08 & 69.38 & 76.48 & 66.02 & 68.40 & 72.62 & 88.02 \\
FastV & 55.12 & 72.31 & \underline{81.92} & 82.16 & 58.73 & 71.53 & 65.61 & 72.18 & 74.56 & 70.46 & 83.13 \\
VisionZip & 64.33 & 77.83 & 80.63 & 87.28 & 67.73 & 73.70 & 75.50 & 69.66 & 76.45 & 74.79 & 88.28 \\
DART & 66.62 & 77.35 & 80.29 & 79.56 & 71.53 & \underline{80.17} & 75.51 & \underline{75.41} & \underline{78.14} & 76.06 & 87.76 \\
PruMerge & 61.62 & 74.19 & 77.14 & 83.27 & 67.65 & 74.05 & 74.09 & 68.88 & 74.91 & 72.87 & 86.74 \\
SparseVLM & \textbf{76.51} & \textbf{84.12} & \textbf{86.11} & \textbf{91.96} & \textbf{82.68} & \textbf{83.01} & \textbf{86.81} & \textbf{86.05} & \textbf{90.12} & \textbf{85.26} & \textbf{93.26} \\
VisPruner & 65.09 & \underline{78.01} & 79.12 & \underline{87.50} & 76.99 & 72.96 & 75.86 & 68.57 & 74.42 & 75.39 & 88.61 \\
\specialrule{0.3pt}{0.4pt}{0pt}
\rowcolor{bandgray}
\multicolumn{12}{c}{\textit{Retain 32 visual tokens}} \\
\specialrule{0pt}{0pt}{0pt}
Random & 55.86 & 54.40 & 54.31 & 52.87 & 57.23 & 55.16 & 56.09 & 54.56 & 54.29 & 54.97 & 73.80 \\
CDPruner & \underline{74.43} & 73.52 & 74.17 & 80.65 & \underline{82.86} & 73.99 & \underline{82.92} & \underline{75.00} & 67.41 & \underline{76.11} & \underline{85.46} \\
DivPrune & 67.54 & 70.52 & 71.81 & 78.18 & 78.02 & 70.57 & 75.64 & 69.76 & 71.48 & 72.61 & 83.90 \\
FastV & 50.54 & 58.26 & 76.67 & 70.51 & 51.05 & 61.92 & 54.83 & 63.60 & 64.89 & 61.36 & 72.54 \\
VisionZip & 59.52 & 73.02 & \underline{78.87} & 81.47 & 67.24 & 73.41 & 70.63 & 71.27 & 71.83 & 71.92 & 82.27 \\
DART & 61.68 & 68.98 & 77.38 & 71.67 & 68.79 & 73.04 & 68.61 & 70.71 & 72.06 & 70.32 & 80.71 \\
PruMerge & 60.00 & \underline{73.68} & 77.13 & 80.99 & 62.37 & \underline{74.49} & 70.37 & 68.58 & \underline{72.98} & 71.18 & 81.92 \\
SparseVLM & \textbf{79.20} & \textbf{82.54} & \textbf{85.69} & \textbf{89.38} & \textbf{86.50} & \textbf{87.40} & \textbf{88.17} & \textbf{88.22} & \textbf{93.60} & \textbf{86.74} & \textbf{91.63} \\
VisPruner & 61.37 & 71.97 & 75.26 & \underline{82.62} & 71.62 & 74.01 & 71.31 & 71.06 & 68.54 & 71.97 & 82.68 \\
\bottomrule[0.55pt]
\end{tabular}
\endgroup

\end{center}
\end{table}

\textbf{SparseVLM holds up best as the budget shrinks.}
SparseVLM has the highest S-Avg and W-Avg at all three budgets. Reducing the budget from 128 to 32 tokens lowers its W-Avg by only 2.80 points, from 94.43 to 91.63. Over the same range, FastV drops by 18.40 points (90.94 to 72.54), and CDPruner drops by 6.64 points (92.10 to 85.46). The methods therefore spread farther apart as compression becomes more aggressive. At 32 tokens, FastV's W-Avg of 72.54 even falls below Random's 73.80.

\textbf{The aggregate does not tell the whole story.}
At 128 tokens, FastV has the highest score on ScienceQA-IMG and MME, while DART leads on MMBench-EN. A method can therefore lead an aggregate without leading every task. The large gap between Random's W-Avg of 85.61 and S-Avg of 56.07 also shows how strongly the sample mix shapes the final score.

\clearpage
\subsection{LLaVA-NeXT-7B}
\label{app:next_results}
\begin{table}[h]
\caption{Complete LLaVA-NeXT results with 640, 320, and 160 retained visual tokens. S-Avg denotes the three-stratum aggregate and W-Avg the sample-weighted aggregate. Bold and underlined values indicate the best and second-best results, respectively, within each column and token budget.}
\label{tab:next_retention_full}
\begin{center}
\definecolor{bandgray}{gray}{0.88}
\begingroup
\small
\newcommand{\paahead}[1]{{\small\bfseries #1}}
\setlength{\tabcolsep}{2.8pt}\setlength{\arrayrulewidth}{0.3pt}
\renewcommand{\arraystretch}{1.0}
\noindent\begin{tabular}{@{}l|cccccccc|cc@{}}
\toprule[0.55pt]
\makebox[44pt][l]{\paahead{Method}} & \paahead{GQA} & \paahead{VizWiz} & \paahead{SQA\textsuperscript{I}} & \paahead{VQA\textsuperscript{T}} & \paahead{POPE} & \paahead{MME} & \paahead{MMB} & \paahead{MMB\textsuperscript{CN}} & \paahead{S-Avg} & \paahead{W-Avg} \\
\specialrule{0.3pt}{0.4pt}{0pt}
\rowcolor{bandgray}
\multicolumn{11}{c}{\textit{Retain 640 visual tokens}} \\
\specialrule{0pt}{0pt}{0pt}
Random & 57.16 & 55.96 & 56.88 & 55.29 & 59.06 & 57.99 & 56.31 & 56.29 & 56.87 & 89.57 \\
CDPruner & 65.81 & 73.74 & 67.49 & 82.59 & 77.47 & 76.54 & 80.35 & 69.28 & 74.16 & 92.87 \\
DivPrune & 62.49 & 69.79 & 68.69 & 75.07 & 74.96 & 75.06 & 67.93 & 58.94 & 69.12 & 91.71 \\
FastV & \underline{79.78} & \underline{84.88} & \underline{84.40} & \textbf{91.48} & 79.95 & \underline{85.02} & \underline{80.84} & \underline{87.45} & \underline{84.22} & 95.76 \\
VisionZip & 64.67 & 75.05 & 63.82 & 86.04 & 73.53 & 72.45 & 73.22 & 66.29 & 71.88 & 92.73 \\
DART & \textbf{83.21} & \textbf{86.82} & \textbf{86.04} & 90.25 & \textbf{89.12} & \textbf{86.33} & 80.67 & \textbf{87.58} & \textbf{86.25} & \textbf{96.13} \\
PruMerge & 60.24 & 66.15 & 59.34 & 71.29 & 72.59 & 65.52 & 50.39 & 57.54 & 62.88 & 89.77 \\
SparseVLM & 76.30 & 82.99 & 75.73 & \underline{90.71} & \underline{82.17} & 77.17 & \textbf{84.02} & 84.44 & 81.69 & \underline{95.84} \\
\specialrule{0.3pt}{0.4pt}{0pt}
\rowcolor{bandgray}
\multicolumn{11}{c}{\textit{Retain 320 visual tokens}} \\
\specialrule{0pt}{0pt}{0pt}
Random & 56.63 & 54.95 & 56.17 & 53.65 & 58.26 & 56.16 & 57.07 & 55.21 & 56.01 & 84.78 \\
CDPruner & 69.90 & 72.49 & 60.90 & 81.73 & \underline{86.47} & 72.73 & 76.37 & 63.10 & 72.96 & 89.82 \\
DivPrune & 63.78 & 69.40 & 62.13 & 74.50 & 80.42 & 70.57 & 73.65 & 64.64 & 69.89 & 88.91 \\
FastV & 72.23 & 77.69 & \textbf{79.35} & \underline{86.40} & 71.01 & 76.12 & 75.91 & 81.31 & 77.50 & 90.84 \\
VisionZip & 63.52 & 75.56 & 61.22 & 85.03 & 75.11 & 68.42 & 70.35 & 63.87 & 70.38 & 88.95 \\
DART & \underline{77.40} & \underline{82.08} & \underline{77.61} & 85.38 & 81.74 & \textbf{82.61} & \underline{79.12} & \underline{85.91} & \underline{81.48} & \underline{92.32} \\
PruMerge & 62.12 & 69.70 & 58.85 & 78.23 & 76.18 & 68.13 & 61.39 & 58.77 & 66.67 & 87.14 \\
SparseVLM & \textbf{78.41} & \textbf{83.60} & 72.57 & \textbf{91.62} & \textbf{89.85} & \underline{81.69} & \textbf{90.07} & \textbf{88.01} & \textbf{84.48} & \textbf{94.56} \\
\specialrule{0.3pt}{0.4pt}{0pt}
\rowcolor{bandgray}
\multicolumn{11}{c}{\textit{Retain 160 visual tokens}} \\
\specialrule{0pt}{0pt}{0pt}
Random & 56.61 & 54.49 & 56.65 & 52.88 & 57.34 & 56.58 & 55.44 & 54.70 & 55.59 & 79.70 \\
CDPruner & \underline{72.30} & 71.74 & 65.88 & 80.95 & \underline{90.56} & 73.16 & \underline{72.20} & 67.79 & \underline{74.32} & \underline{87.68} \\
DivPrune & 64.93 & 70.96 & 67.26 & 75.06 & 81.20 & 67.40 & 71.60 & 67.61 & 70.75 & 86.28 \\
FastV & 60.40 & 61.69 & \textbf{72.79} & 74.03 & 55.82 & 67.83 & 68.36 & 69.87 & 66.35 & 80.69 \\
VisionZip & 60.39 & 72.26 & 67.10 & \underline{81.53} & 68.88 & 61.24 & 66.18 & 61.47 & 67.38 & 83.40 \\
DART & 69.24 & \underline{76.32} & 71.75 & 77.28 & 74.49 & \underline{76.20} & 71.16 & \underline{71.82} & 73.53 & 85.93 \\
PruMerge & 61.28 & 68.12 & 64.76 & 79.99 & 73.09 & 68.72 & 59.87 & 60.81 & 67.08 & 83.08 \\
SparseVLM & \textbf{80.83} & \textbf{83.85} & \underline{72.65} & \textbf{90.95} & \textbf{93.29} & \textbf{86.13} & \textbf{85.93} & \textbf{87.71} & \textbf{85.17} & \textbf{93.18} \\
\bottomrule[0.55pt]
\end{tabular}
\endgroup

\end{center}
\end{table}

\textbf{The ranking changes as the budget tightens.}
At 640 tokens, DART leads both aggregates with an S-Avg/W-Avg of 86.25/96.13. SparseVLM takes the lead at 320 and 160 tokens, reaching 85.17/93.18 at the smallest budget. From 640 to 160 tokens, FastV's W-Avg falls by 15.07 points (95.76 to 80.69), DART's by 10.20 points (96.13 to 85.93), and SparseVLM's by only 2.66 points (95.84 to 93.18). A ranking obtained at a generous budget does not carry over to stronger compression.

\textbf{No single aggregate captures what happens on every task.}
At 160 tokens, SparseVLM leads seven of the eight dataset columns. The only exception is ScienceQA-IMG, where FastV is slightly higher at 72.79 versus 72.65. CDPruner also varies sharply across tasks, scoring 90.56 on POPE but only 65.88 on ScienceQA-IMG. These differences are easy to miss in a single summary number, so the task-level results remain essential.

\textbf{S-Avg makes differences in less common difficulty groups more visible.}
At 640 tokens, DART is 6.56 points above Random in W-Avg but 29.38 points above it in S-Avg. At 160 tokens, SparseVLM's corresponding margins are 13.48 and 29.58 points. The reason is straightforward: S-Avg gives Easy, Medium, and Hard examples equal weight, so groups with fewer samples have more influence than they do under W-Avg.

\clearpage
\subsection{Qwen3-VL-8B}
\begin{table}[htbp]
\caption{Qwen3-VL results at 10\% and 20\% visual-token retention. S-Avg denotes the three-stratum aggregate and W-Avg the sample-weighted aggregate. Bold and underlined values indicate the best and second-best results, respectively, within each column and retention budget.}
\label{tab:retention_results}
\begin{center}
\definecolor{bandgray}{gray}{0.88}
\begingroup
\small
\newcommand{\paahead}[1]{{\small\bfseries #1}}
\setlength{\tabcolsep}{2.8pt}\setlength{\arrayrulewidth}{0.3pt}
\renewcommand{\arraystretch}{1.0}
\noindent\begin{tabular}{@{}l|cccccccc|cc@{}}
\toprule[0.55pt]
\makebox[44pt][l]{\paahead{Method}} & \paahead{GQA} & \paahead{VizWiz} & \paahead{SQA\textsuperscript{I}} & \paahead{VQA\textsuperscript{T}} & \paahead{POPE} & \paahead{MME} & \paahead{MMB} & \paahead{MMB\textsuperscript{CN}} & \paahead{S-Avg} & \paahead{W-Avg} \\
\specialrule{0.3pt}{0.4pt}{0pt}
\rowcolor{bandgray}
\multicolumn{11}{c}{\textit{Retain 10\% of visual tokens}} \\
\specialrule{0pt}{0pt}{0pt}
Random & 56.37 & 55.67 & 55.08 & 54.32 & 56.60 & 56.47 & 56.44 & 56.58 & 55.94 & 79.14 \\
CDPruner & \underline{60.35} & 59.65 & 59.31 & \underline{69.35} & 59.79 & 60.20 & \textbf{64.31} & \textbf{62.14} & 61.89 & \underline{81.62} \\
DivPrune & \textbf{65.50} & \underline{60.55} & 56.87 & \textbf{70.79} & 78.22 & \textbf{63.69} & \underline{61.02} & \underline{57.37} & \textbf{64.25} & \textbf{82.19} \\
FastV & 54.60 & 50.03 & 51.83 & 52.53 & \underline{83.59} & \underline{60.41} & 55.16 & 53.64 & 57.72 & 72.36 \\
VisionZip & 58.23 & \textbf{69.53} & \underline{62.52} & 61.23 & \textbf{85.37} & 60.37 & 57.09 & 54.28 & \underline{63.58} & 80.08 \\
DART & 55.62 & 53.57 & \textbf{66.25} & 50.10 & 66.58 & 56.33 & 54.95 & 57.02 & 57.55 & 76.53 \\
\specialrule{0.3pt}{0.4pt}{0pt}
\rowcolor{bandgray}
\multicolumn{11}{c}{\textit{Retain 20\% of visual tokens}} \\
\specialrule{0pt}{0pt}{0pt}
Random & 56.96 & 56.50 & 55.64 & 55.86 & 58.16 & 57.47 & 58.07 & 57.06 & 56.97 & 86.59 \\
CDPruner & 63.46 & 60.71 & 63.79 & 71.47 & 75.67 & \textbf{67.21} & \textbf{66.79} & \underline{65.06} & \underline{66.77} & 88.53 \\
DivPrune & \textbf{71.02} & 61.66 & 58.59 & \underline{74.17} & 84.21 & 63.76 & 60.68 & 58.14 & 66.53 & \underline{88.95} \\
FastV & 60.73 & \underline{61.96} & 56.25 & 64.04 & \underline{84.79} & 61.84 & 58.07 & 60.24 & 63.49 & 83.29 \\
VisionZip & \underline{66.20} & \textbf{75.10} & \underline{65.36} & \textbf{75.27} & \textbf{88.61} & \underline{66.21} & \underline{64.98} & 62.93 & \textbf{70.58} & \textbf{89.64} \\
DART & 59.62 & 61.90 & \textbf{67.03} & 54.01 & 70.70 & 60.15 & 63.98 & \textbf{66.30} & 62.96 & 84.72 \\
\bottomrule[0.55pt]
\end{tabular}
\endgroup

\end{center}
\end{table}

\textbf{The preferred selector depends on the retention rate.}
At 10\% retention, DivPrune has the highest S-Avg/W-Avg at 64.25/82.19. At 20\%, VisionZip leads with 70.58/89.64. Even at the same 20\% retention rate, CDPruner and DivPrune switch order depending on the aggregate: CDPruner is slightly higher in S-Avg (66.77 versus 66.53), whereas DivPrune is higher in W-Avg (88.95 versus 88.53).

\textbf{A strong result on one task can hide weak results elsewhere.}
With 10\% retention, FastV scores 83.59 on POPE but only 50.03 on VizWiz and 52.53 on TextVQA. Its S-Avg is 1.78 points above Random, yet its W-Avg of 72.36 is 6.78 points below Random. In other words, the same selector can look better or worse depending on how the difficulty groups are weighted. Both aggregates and the per-dataset results are needed before treating an improvement as broad.

\textbf{A strong Random baseline still leaves room for better selection.}
At 20\% retention, Random reaches a W-Avg of 86.59 but an S-Avg of only 56.97. VisionZip raises these scores to 89.64 and 70.58, gains of 3.05 and 13.61 points, respectively. The much larger gain in S-Avg is consistent with the main-paper finding that Easy examples can make improvements look smaller in an aggregate dominated by the observed sample mix.

\clearpage
\subsection{Qwen3-VL-2B}
\label{app:qwen3_2b_results}
\begin{table}[H]
\centering
\caption{Token-selection results for \textbf{Qwen3-VL-2B} on seven benchmarks. Dataset columns report three-stratum preservation scores; S-Avg and W-Avg follow Section~\ref{app:aggregation}. All values are percentages. Bold and underlined values mark the best and second-best scores within each column and budget, respectively.}
\label{tab:qwen3_2b_selection}
\begingroup
\small
\setlength{\tabcolsep}{2.7pt}
\renewcommand{\arraystretch}{1.15}
\definecolor{selectionband}{gray}{0.90}
\begin{tabular*}{\linewidth}{@{\extracolsep{\fill}}l*{9}{r}@{}}
\toprule
\textbf{Method} & \textbf{VizWiz} & \textbf{SQA$^I$} & \textbf{TextVQA} & \textbf{POPE} & \textbf{MME} & \textbf{MMB-EN} & \textbf{MMB-CN} & \textbf{S-Avg} & \textbf{W-Avg} \\
\midrule
\rowcolor{selectionband}
\multicolumn{10}{c}{\strut\textbf{Retain 20\% of visual tokens}} \\
\addlinespace[2pt]
Random & 56.64 & 56.45 & 55.16 & 54.72 & 56.51 & 56.45 & 57.24 & 56.17 & 86.27 \\
CDPruner & 67.43 & \textbf{74.81} & \underline{82.71} & \textbf{100.00} & \textbf{70.51} & \underline{73.08} & \textbf{68.56} & \textbf{76.73} & \underline{91.53} \\
DivPrune & \textbf{68.98} & \underline{71.67} & \textbf{83.34} & 66.67 & \underline{69.58} & \textbf{75.73} & \underline{66.73} & \underline{71.82} & \textbf{92.34} \\
FastV & \underline{67.75} & 49.57 & 72.99 & 44.44 & 61.38 & 53.32 & 53.12 & 57.51 & 83.75 \\
VisionZip & 65.75 & 61.91 & 57.46 & \underline{77.78} & 56.27 & 57.23 & 56.70 & 61.87 & 85.71 \\
DART & 60.25 & 65.75 & 63.09 & 41.66 & 62.18 & 64.18 & 65.82 & 60.42 & 86.64 \\
\addlinespace[4pt]
\midrule
\rowcolor{selectionband}
\multicolumn{10}{c}{\strut\textbf{Retain 10\% of visual tokens}} \\
\addlinespace[2pt]
Random & 55.56 & 55.30 & 53.31 & \underline{58.78} & 55.56 & 56.53 & 55.75 & 55.83 & 79.27 \\
CDPruner & \underline{67.42} & \textbf{73.69} & \underline{78.47} & \textbf{73.33} & \textbf{68.95} & \textbf{69.42} & 65.10 & \textbf{70.91} & \underline{86.56} \\
DivPrune & \textbf{68.34} & \underline{70.08} & \textbf{79.12} & 55.56 & \underline{65.10} & \underline{69.37} & \textbf{65.58} & \underline{67.59} & \textbf{86.86} \\
FastV & 43.69 & 44.89 & 51.10 & 44.44 & 50.91 & 42.80 & 40.95 & 45.54 & 69.27 \\
VisionZip & 54.53 & 52.91 & 44.10 & 56.67 & 52.82 & 53.20 & 50.55 & 52.11 & 75.85 \\
DART & 53.02 & 63.29 & 60.71 & 44.44 & 61.94 & 65.17 & \underline{65.16} & 59.11 & 80.22 \\
\bottomrule
\end{tabular*}
\endgroup

\end{table}

\textbf{S-Avg and W-Avg favor different methods.}
CDPruner has the highest S-Avg at both 20\% and 10\% retention, with scores of 76.73 and 70.91. DivPrune instead has the highest W-Avg, at 92.34 and 86.86. At 20\%, CDPruner is 4.91 points ahead of DivPrune in S-Avg but 0.81 points behind in W-Avg; at 10\%, the corresponding gaps are 3.32 and 0.30 points. Which method leads therefore depends on whether each difficulty group receives equal weight or keeps its observed share of the data.

\textbf{FastV deteriorates sharply under stronger compression.}
When retention drops from 20\% to 10\%, FastV's W-Avg falls from 83.75 to 69.27, whereas Random falls from 86.27 to 79.27. At the smaller budget, FastV is therefore 10.00 points below Random. VisionZip also trails Random at both budgets, while CDPruner and DivPrune remain above Random in both aggregates.

\textbf{A task score only makes sense with its evaluation denominator.}
CDPruner's POPE score of 100.00 at 20\% retention is the three-stratum preservation score computed on examples that the dense model originally answered correctly. It is not 100\% accuracy on the full POPE benchmark. The number of examples in each stratum is also needed to judge the precision and coverage of this result. In addition, this seven-task aggregate excludes GQA and VQAv2, so it should not be compared directly with the aggregate in the Qwen3-VL-8B table.

\clearpage
\subsection{LLaVA-1.5-13B}
\label{app:llava15_13b_results}
\begin{table}[H]
\centering
\caption{Token-selection results for \textbf{LLaVA-1.5-13B} on seven benchmarks. Dataset columns report three-stratum preservation scores; S-Avg and W-Avg follow Section~\ref{app:aggregation}. All values are percentages. Bold and underlined values mark the best and second-best scores within each column and budget, respectively.}
\label{tab:llava15_13b_selection}
\begingroup
\small
\setlength{\tabcolsep}{2.7pt}
\renewcommand{\arraystretch}{1.15}
\definecolor{selectionband}{gray}{0.90}
\begin{tabular*}{\linewidth}{@{\extracolsep{\fill}}l*{9}{r}@{}}
\toprule
\textbf{Method} & \textbf{VizWiz} & \textbf{SQA$^I$} & \textbf{TextVQA} & \textbf{POPE} & \textbf{MME} & \textbf{MMB-EN} & \textbf{MMB-CN} & \textbf{S-Avg} & \textbf{W-Avg} \\
\midrule
\rowcolor{selectionband}
\multicolumn{10}{c}{\strut\textbf{Retain 128 visual tokens}} \\
\addlinespace[2pt]
Random & 56.44 & 58.24 & 54.60 & 57.23 & 58.04 & 56.39 & 55.00 & 56.56 & 86.34 \\
CDPruner & 73.72 & 73.08 & 85.44 & 66.80 & 69.79 & 71.35 & 67.20 & 72.48 & 92.44 \\
DivPrune & 74.54 & 68.76 & 81.40 & 64.78 & 66.65 & 67.71 & 63.49 & 69.62 & 91.88 \\
FastV & \textbf{82.73} & \textbf{88.90} & \underline{86.38} & \underline{79.88} & \textbf{89.36} & \textbf{84.87} & \underline{78.63} & \textbf{84.39} & \underline{93.95} \\
VisionZip & 78.82 & 78.40 & \textbf{88.11} & 68.05 & 70.17 & 69.59 & 67.77 & 74.42 & 93.07 \\
DART & \underline{82.60} & \underline{88.89} & 85.98 & \textbf{82.46} & \underline{86.56} & \underline{79.32} & \textbf{84.70} & \underline{84.36} & \textbf{94.25} \\
\addlinespace[4pt]
\midrule
\rowcolor{selectionband}
\multicolumn{10}{c}{\strut\textbf{Retain 64 visual tokens}} \\
\addlinespace[2pt]
Random & 55.26 & 55.00 & 53.46 & 57.41 & 57.87 & 55.91 & 55.26 & 55.74 & 81.28 \\
CDPruner & \underline{75.16} & 72.21 & \underline{84.11} & 73.17 & 71.47 & 75.36 & 71.34 & 74.69 & \underline{90.13} \\
DivPrune & 74.01 & 66.07 & 79.71 & 70.59 & 71.91 & 70.10 & 67.72 & 71.44 & 88.98 \\
FastV & 74.73 & \textbf{83.20} & 78.31 & \underline{73.76} & \underline{78.80} & \underline{76.00} & \underline{79.80} & \underline{77.80} & 88.04 \\
VisionZip & \textbf{78.16} & 80.14 & \textbf{86.59} & 69.05 & 72.38 & 74.20 & 72.07 & 76.09 & \textbf{90.37} \\
DART & 73.72 & \underline{80.26} & 75.98 & \textbf{76.25} & \textbf{80.00} & \textbf{82.09} & \textbf{81.79} & \textbf{78.59} & 89.01 \\
\addlinespace[4pt]
\midrule
\rowcolor{selectionband}
\multicolumn{10}{c}{\strut\textbf{Retain 32 visual tokens}} \\
\addlinespace[2pt]
Random & 54.17 & 55.38 & 52.66 & 56.52 & 56.59 & 55.36 & 55.14 & 55.12 & 75.32 \\
CDPruner & \underline{73.83} & 74.59 & \underline{78.11} & \textbf{78.69} & \underline{74.75} & \textbf{74.07} & 71.97 & \textbf{75.15} & \textbf{86.26} \\
DivPrune & 71.80 & 71.58 & 76.05 & \underline{74.93} & 74.14 & 66.92 & 73.50 & 72.70 & \underline{85.27} \\
FastV & 63.41 & \textbf{81.40} & 67.46 & 69.46 & 69.54 & 66.88 & 70.60 & 69.82 & 79.84 \\
VisionZip & \textbf{74.34} & 77.85 & \textbf{81.42} & 67.30 & 67.87 & \underline{71.77} & \underline{74.35} & \underline{73.56} & 85.07 \\
DART & 63.10 & \underline{79.28} & 68.01 & 73.93 & \textbf{75.73} & 71.38 & \textbf{74.37} & 72.26 & 81.79 \\
\bottomrule
\end{tabular*}
\endgroup

\end{table}

\textbf{The leader changes with both the budget and the aggregate.}
At 128 tokens, FastV has the highest S-Avg at 84.39, only 0.03 points above DART, while DART has the highest W-Avg at 94.25. At 64 tokens, DART leads S-Avg (78.59) and VisionZip leads W-Avg (90.37). At 32 tokens, CDPruner leads both aggregates, with scores of 75.15 and 86.26. The ordering at a larger budget therefore does not tell us what the ordering will be under stronger compression.

\textbf{Some task-specific strengths persist across budgets.}
VisionZip has the highest TextVQA score at all three budgets---88.11, 86.59, and 81.42---even when another method leads the aggregate. From 128 to 32 tokens, CDPruner's W-Avg drops by 6.18 points, compared with 14.11 points for FastV and 12.46 points for DART. This separates two questions that an overall score can blur: which method is strongest on a particular task, and which method remains stable as the token budget shrinks. As with the other backbones, changes in S-Avg also partly reflect the fact that the difficulty strata themselves change with the budget.

\subsection{Reading the Results across Model Sizes}
Taken together, the five tables show that method rankings change with the budget, the task, and the aggregation rule. 
Fixed-budget preservation also says nothing about realized efficiency. The next section therefore reports measured speedups only for the three primary model configurations.
\clearpage
\section{Preservation-Constrained Speedup Results and Analysis}
\label{app:pcs_complete}
The following three tables report the updated operating points for three model backbones on MMBench-EN, TextVQA, POPE, and GQA. Each backbone has a separate table, with results shown at 95\% and 90\% correctness preservation. These percentages refer to the fraction of predictions that were correct under the dense model and remain correct after pruning; they do not indicate the fraction of visual tokens retained. Speedup is measured relative to the corresponding dense-model baseline, while the actual number of retained tokens is reported separately.

\subsection{LLaVA-1.5-7B}
\begin{table}[H]
\centering
\caption{End-to-end speedup and retained visual tokens for \textbf{LLaVA-1.5-7B}. Bold marks the highest reported speedup within each dataset and preservation target, including ties and Random. Dashes denote unreported results.}
\label{tab:pcs_llava15}
\label{tab:pcs_complete}
\begingroup
\small
\setlength{\tabcolsep}{3.5pt}
\renewcommand{\arraystretch}{1.18}
\definecolor{pcstarget}{gray}{0.90}
\begin{tabular*}{\linewidth}{@{\extracolsep{\fill}}l*{4}{rr}@{}}
\toprule
\makebox[49pt][l]{\textbf{Method}} & \multicolumn{2}{c}{\textbf{MMBench-EN}} & \multicolumn{2}{c}{\textbf{TextVQA}} & \multicolumn{2}{c}{\textbf{POPE}} & \multicolumn{2}{c}{\textbf{GQA}} \\
\cmidrule(lr){2-3}\cmidrule(lr){4-5}\cmidrule(lr){6-7}\cmidrule(l){8-9}
& \makebox[32pt][r]{Speedup} & \makebox[28pt][r]{Tokens} & \makebox[32pt][r]{Speedup} & \makebox[28pt][r]{Tokens} & \makebox[32pt][r]{Speedup} & \makebox[28pt][r]{Tokens} & \makebox[32pt][r]{Speedup} & \makebox[28pt][r]{Tokens} \\
\midrule
\rowcolor{pcstarget}
\multicolumn{9}{c}{\strut\textbf{95\% correctness-preservation target}} \\
\addlinespace[2pt]
Dense & 1.00$\times$ & 576 & 1.00$\times$ & 576 & 1.00$\times$ & 576 & 1.00$\times$ & 576 \\
\addlinespace[2pt]
Random & 1.41$\times$ & 266 & 1.08$\times$ & 461 & 1.71$\times$ & 177 & 1.19$\times$ & 426 \\
VisionZip & 1.56$\times$ & 183 & 1.32$\times$ & 230 & 1.70$\times$ & 145 & 1.18$\times$ & 392 \\
FastV & 1.71$\times$ & 115 & \textbf{1.41}$\times$ & 142 & 1.33$\times$ & 270 & 1.23$\times$ & 350 \\
SparseVLM & \textbf{1.74}$\times$ & 26 & 1.33$\times$ & 116 & 1.74$\times$ & 29 & 1.22$\times$ & 274 \\
CDPruner & 1.35$\times$ & 156 & 1.04$\times$ & 262 & 1.98$\times$ & 16 & 0.90$\times$ & 332 \\
DivPrune & 1.53$\times$ & 175 & 1.20$\times$ & 297 & \textbf{2.08}$\times$ & 52 & 1.18$\times$ & 347 \\
DART & 1.72$\times$ & 100 & 1.33$\times$ & 180 & 1.59$\times$ & 180 & \textbf{1.26}$\times$ & 308 \\
\addlinespace[4pt]
\midrule
\rowcolor{pcstarget}
\multicolumn{9}{c}{\strut\textbf{90\% correctness-preservation target}} \\
\addlinespace[2pt]
Dense & 1.00$\times$ & 576 & 1.00$\times$ & 576 & 1.00$\times$ & 576 & 1.00$\times$ & 576 \\
\addlinespace[2pt]
Random & 1.82$\times$ & 107 & 1.24$\times$ & 312 & 1.89$\times$ & 81 & 1.49$\times$ & 216 \\
VisionZip & 1.94$\times$ & 46 & \textbf{1.53}$\times$ & 85 & 1.84$\times$ & 73 & 1.46$\times$ & 201 \\
FastV & 1.83$\times$ & 58 & 1.51$\times$ & 62 & 1.64$\times$ & 172 & 1.42$\times$ & 243 \\
SparseVLM & 1.83$\times$ & 4 & 1.45$\times$ & 23 & 1.84$\times$ & 5 & 1.52$\times$ & 86 \\
CDPruner & 1.91$\times$ & 33 & 1.35$\times$ & 94 & 2.03$\times$ & 6 & 1.42$\times$ & 115 \\
DivPrune & \textbf{1.95}$\times$ & 49 & \textbf{1.53}$\times$ & 85 & \textbf{2.13}$\times$ & 25 & \textbf{1.61}$\times$ & 139 \\
DART & \multicolumn{2}{c}{---} & 1.47$\times$ & 76 & 1.75$\times$ & 84 & 1.58$\times$ & 148 \\
\bottomrule
\end{tabular*}
\endgroup

\end{table}

\paragraph{The leading method depends on the task and preservation target.}
At 95\% correctness preservation, SparseVLM achieves the highest speedup on MMBench-EN ($1.74\times$), FastV on TextVQA ($1.41\times$), DivPrune on POPE ($2.08\times$), and DART on GQA ($1.26\times$). When the preservation target is relaxed to 90\%, DivPrune reaches the highest speedup on all four tasks, tying VisionZip at $1.53\times$ on TextVQA. Its GQA speedup also increases from $1.18\times$ to $1.61\times$, surpassing DART's $1.58\times$. Thus, the ordering of methods at one preservation target does not determine their ordering at another.

\paragraph{Fewer tokens do not necessarily yield greater speedup.}
On MMBench-EN at the 90\% preservation target, SparseVLM retains only 4 tokens and reaches a $1.853imes$ speedup, whereas DivPrune retains 49 tokens but is faster at $1.95\times$. On GQA at 95\% preservation, CDPruner retains 332 of the dense model's 576 tokens yet achieves only $0.90\times$, meaning that pruning causes a net slowdown at this operating point. The final token count does not capture computation already performed before pruning, nor does it reflect the extra overhead introduced by token selection itself.

\clearpage
\subsection{LLaVA-NeXT-7B}
\begin{table}[H]
\centering
\caption{End-to-end speedup and retained visual tokens for \textbf{LLaVA-NeXT-7B}. Bold marks the highest reported speedup within each dataset and preservation target, including ties and Random. Dashes denote unreported results.}
\label{tab:pcs_next}
\begingroup
\small
\setlength{\tabcolsep}{3.5pt}
\renewcommand{\arraystretch}{1.18}
\definecolor{pcstarget}{gray}{0.90}
\begin{tabular*}{\linewidth}{@{\extracolsep{\fill}}l*{4}{rr}@{}}
\toprule
\makebox[49pt][l]{\textbf{Method}} & \multicolumn{2}{c}{\textbf{MMBench-EN}} & \multicolumn{2}{c}{\textbf{TextVQA}} & \multicolumn{2}{c}{\textbf{POPE}} & \multicolumn{2}{c}{\textbf{GQA}} \\
\cmidrule(lr){2-3}\cmidrule(lr){4-5}\cmidrule(lr){6-7}\cmidrule(l){8-9}
& \makebox[32pt][r]{Speedup} & \makebox[28pt][r]{Tokens} & \makebox[32pt][r]{Speedup} & \makebox[28pt][r]{Tokens} & \makebox[32pt][r]{Speedup} & \makebox[28pt][r]{Tokens} & \makebox[32pt][r]{Speedup} & \makebox[28pt][r]{Tokens} \\
\midrule
\rowcolor{pcstarget}
\multicolumn{9}{c}{\strut\textbf{95\% correctness-preservation target}} \\
\addlinespace[2pt]
Dense & 1.00$\times$ & 1927 & 1.00$\times$ & 2310 & 1.00$\times$ & 2250 & 1.00$\times$ & 2239 \\
\addlinespace[2pt]
Random & 2.07$\times$ & 617 & 1.12$\times$ & 1951 & 2.43$\times$ & 571 & 1.35$\times$ & 1482 \\
VisionZip & 1.94$\times$ & 626 & 1.07$\times$ & 2022 & 2.47$\times$ & 446 & \multicolumn{2}{c}{---} \\
FastV & 2.24$\times$ & 385 & \textbf{1.84}$\times$ & 577 & 2.29$\times$ & 474 & 1.73$\times$ & 896 \\
SparseVLM & \textbf{2.53}$\times$ & 65 & 1.59$\times$ & 623 & 2.80$\times$ & 40 & 1.54$\times$ & 906 \\
CDPruner & 2.13$\times$ & 474 & \multicolumn{2}{c}{---} & \textbf{4.37}$\times$ & 44 & \multicolumn{2}{c}{---} \\
DivPrune & 2.01$\times$ & 525 & 0.81$\times$ & 2199 & 3.11$\times$ & 251 & \multicolumn{2}{c}{---} \\
DART & 2.35$\times$ & 355 & 1.81$\times$ & 579 & 2.71$\times$ & 281 & \textbf{1.82}$\times$ & 783 \\
\addlinespace[4pt]
\midrule
\rowcolor{pcstarget}
\multicolumn{9}{c}{\strut\textbf{90\% correctness-preservation target}} \\
\addlinespace[2pt]
Dense & 1.00$\times$ & 1927 & 1.00$\times$ & 2310 & 1.00$\times$ & 2250 & 1.00$\times$ & 2239 \\
\addlinespace[2pt]
Random & 3.31$\times$ & 174 & 1.30$\times$ & 1478 & 3.20$\times$ & 260 & 2.25$\times$ & 653 \\
VisionZip & 3.16$\times$ & 157 & 2.08$\times$ & 468 & 3.26$\times$ & 196 & 1.80$\times$ & 891 \\
FastV & 2.75$\times$ & 177 & 2.15$\times$ & 276 & 2.67$\times$ & 280 & 2.29$\times$ & 448 \\
SparseVLM & 2.79$\times$ & 9 & 1.96$\times$ & 138 & 2.92$\times$ & 11 & 2.23$\times$ & 233 \\
CDPruner & \textbf{3.76}$\times$ & 45 & 1.69$\times$ & 717 & \textbf{4.62}$\times$ & 10 & 2.22$\times$ & 558 \\
DivPrune & 3.68$\times$ & 76 & 1.27$\times$ & 1155 & 3.74$\times$ & 97 & 1.87$\times$ & 705 \\
DART & 2.93$\times$ & 140 & \textbf{2.17}$\times$ & 288 & 3.15$\times$ & 156 & \textbf{2.48}$\times$ & 371 \\
\bottomrule
\end{tabular*}
\endgroup

\end{table}

\paragraph{The attainable speedup remains task dependent.}
At 95\% correctness preservation, the highest reported speedup ranges from DART's $1.82\times$ on GQA to CDPruner's $4.37\times$ on POPE. SparseVLM is fastest on MMBench-EN at $2.53\times$, while FastV leads on TextVQA at $1.84\times$. On POPE, CDPruner retains 44 tokens and is still faster than SparseVLM, which retains 40 tokens and reaches $2.80\times$. This again shows that similar final token budgets can lead to different realized runtime costs across methods.

\paragraph{Relaxing the preservation target to 90\% benefits methods differently.}
CDPruner reaches $3.76\times$ on MMBench-EN and $4.62\times$ on POPE, while DART leads on TextVQA and GQA at $2.17\times$ and $2.48\times$, respectively. CDPruner's newly reported POPE operating point retains only 10 tokens and improves further over its $4.37\times$ result at 95\% preservation. On TextVQA, DivPrune improves from $0.81\times$ at 95\% preservation to $1.27\times$ at 90\%. This also shows that satisfying a correctness-preservation constraint does not necessarily translate into a net speedup.

\paragraph{Task-level differences matter when comparing speedups.}
The dense-model token count is not the same across tasks: MMBench-EN starts from 1,927 tokens, while TextVQA starts from 2,310. The four tasks therefore do not have identical initial workloads. The relatively modest speedups on TextVQA are consistent with the task's sensitivity to fine-grained visual information, but this table alone cannot separate that effect from image resolution, generation length, pruning location, or implementation details. At 90\% preservation, the gap between DART and FastV on TextVQA is also small ($2.17\times$ versus $2.15\times$).

\clearpage
\subsection{Qwen3-VL-8B}
\begin{table}[H]
\centering
\caption{End-to-end speedup and retained visual tokens for \textbf{Qwen3-VL-8B}. Bold marks the highest reported speedup within each dataset and preservation target, including ties and Random. Dashes denote unreported results.}
\label{tab:pcs_qwen}
\begingroup
\small
\setlength{\tabcolsep}{3.5pt}
\renewcommand{\arraystretch}{1.18}
\definecolor{pcstarget}{gray}{0.90}
\begin{tabular*}{\linewidth}{@{\extracolsep{\fill}}l*{4}{rr}@{}}
\toprule
\makebox[49pt][l]{\textbf{Method}} & \multicolumn{2}{c}{\textbf{MMBench-EN}} & \multicolumn{2}{c}{\textbf{TextVQA}} & \multicolumn{2}{c}{\textbf{POPE}} & \multicolumn{2}{c}{\textbf{GQA}} \\
\cmidrule(lr){2-3}\cmidrule(lr){4-5}\cmidrule(lr){6-7}\cmidrule(l){8-9}
& \makebox[32pt][r]{Speedup} & \makebox[28pt][r]{Tokens} & \makebox[32pt][r]{Speedup} & \makebox[28pt][r]{Tokens} & \makebox[32pt][r]{Speedup} & \makebox[28pt][r]{Tokens} & \makebox[32pt][r]{Speedup} & \makebox[28pt][r]{Tokens} \\
\midrule
\rowcolor{pcstarget}
\multicolumn{9}{c}{\strut\textbf{95\% correctness-preservation target}} \\
\addlinespace[2pt]
Dense & 1.00$\times$ & 155 & 1.00$\times$ & 750 & 1.00$\times$ & 269 & 1.00$\times$ & 268 \\
\addlinespace[2pt]
Random & 1.12$\times$ & 45 & 1.11$\times$ & 489 & 1.26$\times$ & 77 & 1.17$\times$ & 186 \\
VisionZip & 1.02$\times$ & 45 & 0.95$\times$ & 289 & 1.10$\times$ & 56 & 0.97$\times$ & 141 \\
FastV & 1.03$\times$ & 65 & 1.15$\times$ & 423 & 1.22$\times$ & 99 & 1.16$\times$ & 179 \\
SparseVLM & 1.12$\times$ & 17 & \textbf{1.23}$\times$ & 167 & 1.31$\times$ & 30 & 1.17$\times$ & 109 \\
CDPruner & 1.13$\times$ & 45 & 0.97$\times$ & 409 & 1.35$\times$ & 47 & 1.03$\times$ & 165 \\
DivPrune & \textbf{1.15}$\times$ & 45 & 1.11$\times$ & 360 & \textbf{1.47}$\times$ & 43 & \textbf{1.18}$\times$ & 151 \\
DART & 1.10$\times$ & 48 & 1.09$\times$ & 480 & 1.21$\times$ & 77 & 1.14$\times$ & 153 \\
\addlinespace[4pt]
\midrule
\rowcolor{pcstarget}
\multicolumn{9}{c}{\strut\textbf{90\% correctness-preservation target}} \\
\addlinespace[2pt]
Dense & 1.00$\times$ & 155 & 1.00$\times$ & 750 & 1.00$\times$ & 269 & 1.00$\times$ & 268 \\
\addlinespace[2pt]
Random & 1.13$\times$ & 24 & 1.19$\times$ & 310 & \textbf{1.45}$\times$ & 34 & 1.23$\times$ & 107 \\
VisionZip & 1.03$\times$ & 25 & 1.02$\times$ & 180 & 1.10$\times$ & 32 & 1.02$\times$ & 84 \\
FastV & 1.04$\times$ & 36 & 1.21$\times$ & 291 & 1.38$\times$ & 53 & 1.20$\times$ & 122 \\
SparseVLM & 1.14$\times$ & 4 & \textbf{1.31}$\times$ & 76 & 1.32$\times$ & 9 & \textbf{1.25}$\times$ & 56 \\
CDPruner & 1.14$\times$ & 19 & 1.08$\times$ & 261 & 1.37$\times$ & 26 & 1.11$\times$ & 91 \\
DivPrune & \textbf{1.18}$\times$ & 20 & 1.25$\times$ & 216 & 1.51$\times$ & 24 & \textbf{1.25}$\times$ & 72 \\
DART & 1.17$\times$ & 25 & 1.14$\times$ & 346 & 1.35$\times$ & 39 & 1.18$\times$ & 103 \\
\bottomrule
\end{tabular*}
\endgroup

\end{table}

\paragraph{Selection gains remain task specific, and Random can be competitive.}
At 95\% correctness preservation, DivPrune leads on MMBench-EN, POPE, and GQA with speedups of $1.15\times$, $1.47\times$, and $1.18\times$, respectively, while SparseVLM leads on TextVQA at $1.23\times$. At 90\% preservation, SparseVLM remains the fastest method on TextVQA ($1.31\times$) and ties DivPrune on GQA at $1.25\times$. In this block, Random achieves the highest reported speedup on POPE at $1.45\times$.

\paragraph{Stronger compression alone does not guarantee a speed benefit.}
At 95\% preservation, VisionZip retains 289 of 750 tokens on TextVQA but reaches only $0.95\times$ speedup; on GQA, it retains 141 of 268 tokens and reaches only $0.97\times$. CDPruner is also slower than the dense model on TextVQA, with a speedup of $0.97\times$. On MMBench-EN, SparseVLM retains fewer tokens than DivPrune at both preservation targets, yet DivPrune is faster in both blocks. These results show why efficiency should be compared using measured inference time under the same correctness-preservation requirement, rather than inferred from the final number of retained tokens.

\clearpage

\clearpage
\section{Qualitative Analysis of Random-Pruning Sensitivity}
\label{app:qualitative}

\subsection{Case Selection, Scoring, and Visual Encoding}
\begin{figure}[H]
\centering
\includegraphics[width=\linewidth]{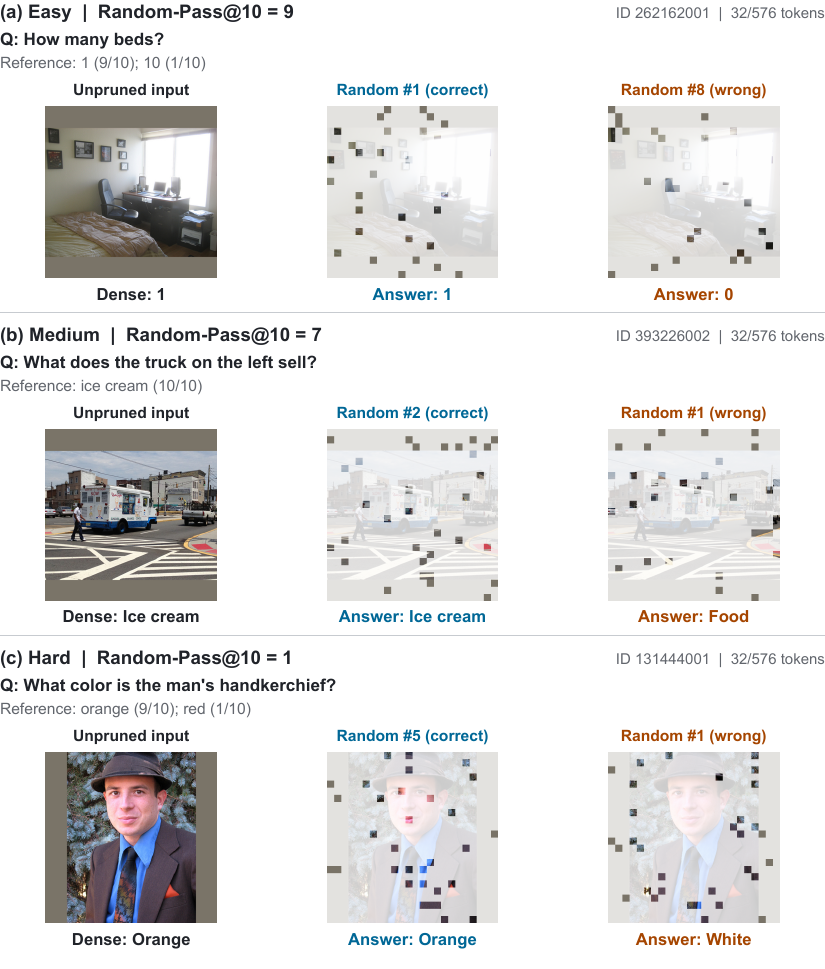}
\caption{\textbf{Random pruning on LLaVA-1.5-7B at 32/576 tokens.} Each VQAv2 row pairs the unpruned model input with one successful and one unsuccessful saved mask at the same token budget. Blue and orange answer labels also indicate in text whether the prediction is correct. Clear patches are retained, while faded patches are removed. Trial numbers are one-based; all ten outcomes are reported in Table~\ref{tab:app_qualitative_trials}.}
\label{fig:app_random_llava15}
\end{figure}

\begin{figure}[H]
\centering
\includegraphics[width=\linewidth]{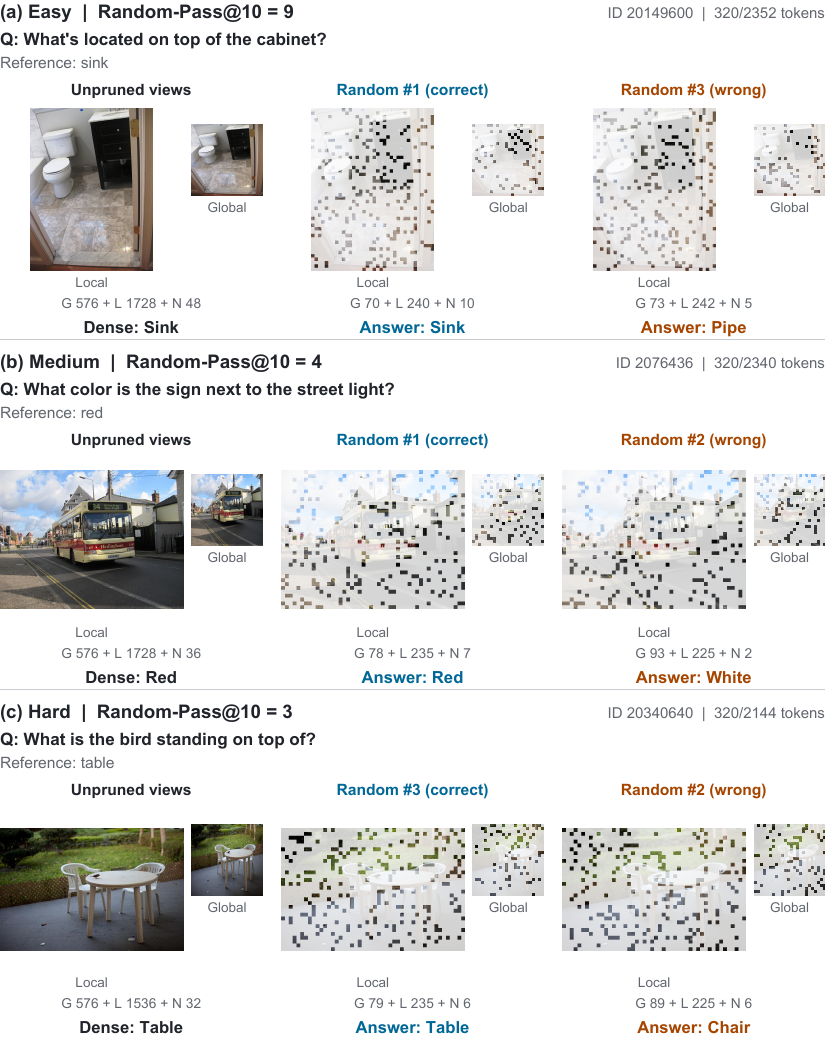}
\caption{\textbf{Random pruning on LLaVA-NeXT-7B at 320 tokens.} The GQA examples show separate local and global views for the dense input and for both selected masks. G, L, and N denote the numbers of global, local, and image-newline tokens, respectively; together they sum to the corresponding token budget. Newline tokens do not correspond to spatial image patches. Clear positions are retained, while faded positions are removed.}
\label{fig:app_random_next}
\end{figure}

\begin{figure}[H]
\centering
\includegraphics[width=\linewidth]{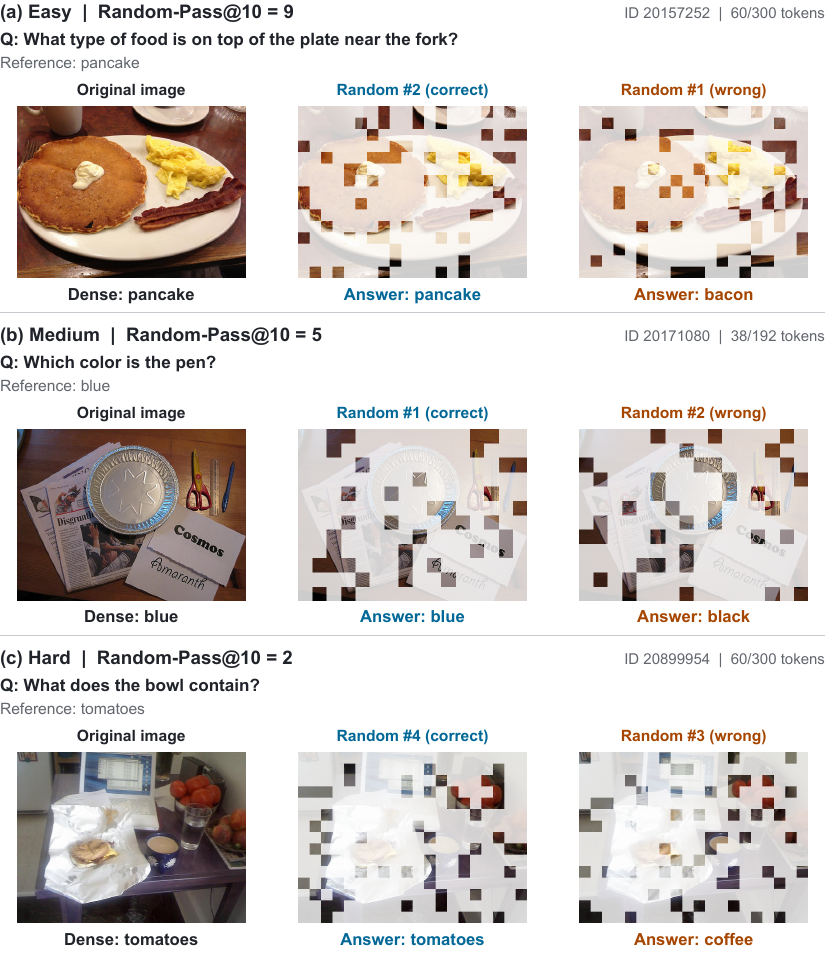}
\caption{\textbf{Random pruning on Qwen3-VL-8B at a nominal 20\% token budget.} The GQA examples pair the original image with saved overlays from one successful and one unsuccessful token-selection trial. The actual budgets are 60/300, 38/192, and 60/300 tokens; the middle case uses $\lfloor0.2\times192\rfloor=38$. Clear positions are retained, while faded positions are removed. The original image is included only as a spatial reference and should not be read as a visualization of internal model features.}
\label{fig:app_random_qwen}
\end{figure}

Figures~\ref{fig:app_random_llava15}--\ref{fig:app_random_qwen} show nine examples that the dense model answers correctly: one Easy, one Medium, and one Hard case for each of the three backbones. The LLaVA-1.5-7B examples come from VQAv2 val2014, while the LLaVA-NeXT-v1.6-Vicuna-7B and Qwen3-VL-8B examples come from GQA testdev-balanced. All images, masks, and predictions are taken from saved experimental records; we do not run any additional inference for this analysis. These cases are meant to illustrate the behavior we observe, rather than serve as a random sample for estimating the prevalence of each difficulty group or comparing backbones. Each case includes one successful and one unsuccessful random-pruning trial. As a result, examples with zero or ten successes do not appear in this paired display.

\paragraph{Scoring and traceability.}
Random-Pass@10 simply counts how many of ten random-pruning trials remain correct: 8--10 successes are labeled Easy, 4--7 Medium, and 0--3 Hard. For the VQAv2 cases shown here, we use the recorded consensus-score-positive rule; for GQA, we use normalized exact matching. All displayed successful answers have score 1, and all failures have score 0. Each figure row lists the sample, the retained and dense token counts, the reference answer, the dense-model prediction, and the trial numbers being shown. For VQAv2, we keep the original reference-answer frequencies rather than collapsing them into a single label. Table~\ref{tab:app_qualitative_trials} reports the predictions from all ten trials for every case, including trials whose masks are not shown in the figures.

\begin{table}[H]
\centering
\caption{\textbf{Complete random-trial answers for the nine displayed cases.} Columns index trials 1--10. Bold answers are correct under the saved scorer; plain answers are incorrect. Cases appear in the same order as the figure rows. Each dense prediction is correct.}
\label{tab:app_qualitative_trials}
\begingroup
\scriptsize
\setlength{\tabcolsep}{2.2pt}
\renewcommand{\arraystretch}{1.12}
\begin{tabular}{@{}l*{10}{c}@{}}
\toprule
Case & 1 & 2 & 3 & 4 & 5 & 6 & 7 & 8 & 9 & 10 \\
\midrule
\multicolumn{11}{@{}l}{\textit{LLaVA-1.5-7B / VQAv2 / 32 tokens}} \\
Easy & \textbf{1} & \textbf{1} & \textbf{1} & \textbf{1} & \textbf{1} & \textbf{1} & \textbf{1} & 0 & \textbf{1} & \textbf{1} \\
Medium & Food & \textbf{Ice cream} & \textbf{Ice cream} & \textbf{Ice cream} & \textbf{Ice cream} & Food & \textbf{Ice cream} & \textbf{Ice cream} & \textbf{Ice cream} & Coffee \\
Hard & White & White & White & White & \textbf{Orange} & Blue & Brown & White & Blue & White \\
\midrule
\multicolumn{11}{@{}l}{\textit{LLaVA-NeXT-7B / GQA / 320 tokens}} \\
Easy & \textbf{Sink} & \textbf{Sink} & Pipe & \textbf{Sink} & \textbf{Sink} & \textbf{Sink} & \textbf{Sink} & \textbf{Sink} & \textbf{Sink} & \textbf{Sink} \\
Medium & \textbf{Red} & White & White & Black & Yellow & Yellow & \textbf{Red} & White & \textbf{Red} & \textbf{Red} \\
Hard & Fence & Chair & \textbf{Table} & \textbf{Table} & \textbf{Table} & Chair & Chair & Fence & Chair & Chair \\
\midrule
\multicolumn{11}{@{}l}{\textit{Qwen3-VL-8B / GQA / 20\% retained}} \\
Easy & bacon & \textbf{pancake} & \textbf{pancake} & \textbf{pancake} & \textbf{pancake} & \textbf{pancake} & \textbf{pancake} & \textbf{pancake} & \textbf{pancake} & \textbf{pancake} \\
Medium & \textbf{blue} & black & \textbf{blue} & \textbf{blue} & yellow & \textbf{blue} & \textbf{blue} & red & black & yellow \\
Hard & fruit & peaches & coffee & \textbf{tomatoes} & \textbf{tomatoes} & apples & peaches & coffee & coffee & coffee \\
\bottomrule
\end{tabular}
\endgroup
\end{table}

\paragraph{Spatial encoding and its limits.}
Clear patches indicate retained token positions, while removed positions are covered by an 80\% white overlay. These overlays show token selection after visual encoding, rather than pixel masking before the vision encoder. The LLaVA-1.5 visualizations keep the model's square padding. For LLaVA-NeXT, the global view and the high-resolution local view are shown separately; image-newline tokens count toward the token budget but do not correspond to image patches. The Qwen overlays use the saved spatial mapping, with the retained count rounded down from 20\% of the native visual-token count. Matching views within the same case use the same scale, and the original image aspect ratio is preserved.

\clearpage
\subsection{LLaVA-1.5-7B: Counting, Recognition, and Local Attributes}
\noindent\textbf{Reading the cases.} The bed-counting question remains correct in nine of the ten trials, but one mask changes the answer from ``1'' to ``0''. The truck-recognition question succeeds in seven trials; when it fails, the model answers ``Food'' or ``Coffee'' instead of ``Ice cream''. The handkerchief-color question succeeds only once, and ``White'' is the most common incorrect answer. These examples show that the three questions differ substantially in how sensitive they are to random token selection.

\subsection{LLaVA-NeXT-7B: Global and Local Token Views}

\clearpage
\subsection{Qwen3-VL-8B: Relative Budgets and Answer Variability}
\noindent\textbf{Reading the cases.} The pancake question fails only once in ten trials, producing ``bacon''. The pen-color question produces ``blue'' in five trials, with the remaining answers split among black, yellow, and red. The bowl-content question produces ``tomatoes'' only twice, while ``coffee'' is the most common incorrect answer. 

\clearpage
\clearpage
\section{Additional Analysis of Random-Pass Stratification}
\label{app:random-pass-analysis}

We examine how Random-Pass stratification varies across datasets and
token budgets, and how its results compare when more random-pruning
trials are used.

\subsection{Difficulty Distributions across Token Budgets}
\label{app:dataset-difficulty-distributions}

Figures~\ref{fig:difficulty-pies-128}--\ref{fig:difficulty-pies-32}
show the difficulty composition of dense-correct examples for LLaVA-1.5
on eight datasets at 128, 64, and 32 retained visual tokens.
At each budget, Random-Pass@10 assigns examples to Easy (8--10 correct
trials), Medium (4--7), or Hard (0--3). Smaller budgets increase the Hard
share on every dataset, but the shift is dataset-dependent: TextVQA's
Hard share grows from 31.80\% to 61.30\%, whereas ScienceQA-IMG remains
82.58\% Easy at 32 tokens. These differences motivate reporting
difficulty-stratified results alongside aggregate performance.

\noindent\begin{minipage}{\linewidth}
  \captionsetup{type=figure,hypcap=false}
  \centering
  \includegraphics[width=\linewidth]{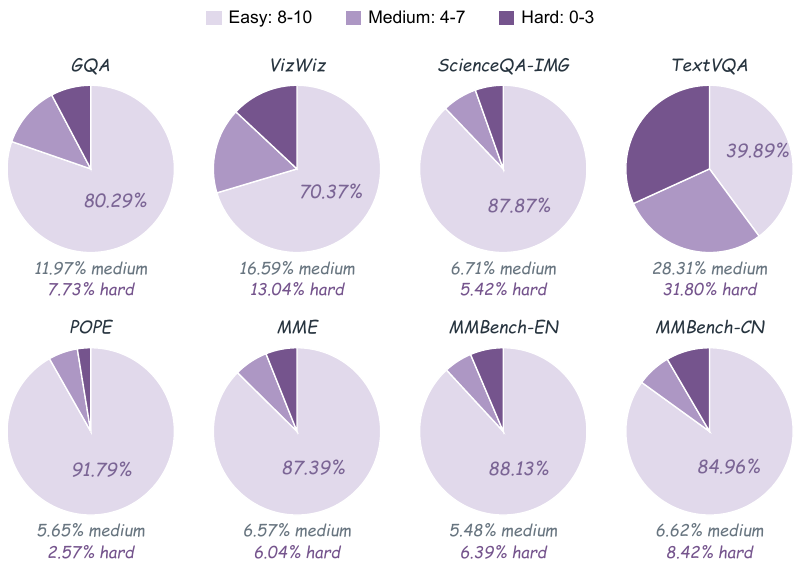}

  \caption{\textbf{Selection-difficulty distributions at 128 retained visual tokens.}
  Each pie reports the Easy, Medium, and Hard shares of dense-correct examples
  for LLaVA-1.5 on one dataset, using Random-Pass@10.
  The percentage inside each pie denotes Easy; the two values below denote
  Medium and Hard. Percentages may not sum to exactly 100\% due to rounding.}
  \label{fig:difficulty-pies-128}
\end{minipage}

\clearpage

\noindent\begin{minipage}{\linewidth}
  \captionsetup{type=figure,hypcap=false}
  \centering
  \includegraphics[width=0.95\linewidth]{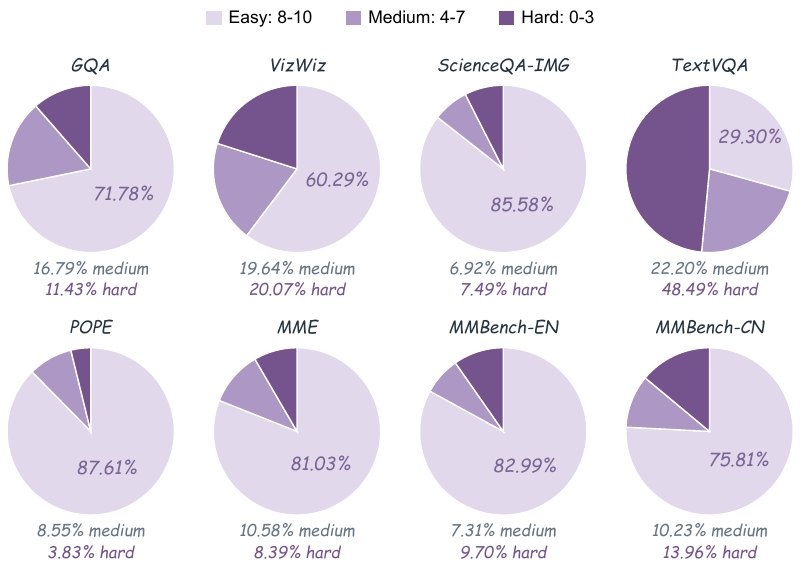}

  \caption{\textbf{Selection-difficulty distributions at 64 retained visual tokens.}
  Dataset order, colors, and annotations follow Figure~\ref{fig:difficulty-pies-128}.
  Relative to 128 tokens, every dataset has a smaller Easy share and a
  larger Hard share; TextVQA has the largest Hard share (48.49\%).}
  \label{fig:difficulty-pies-64}
\end{minipage}

\vspace{10pt}
\noindent\begin{minipage}{\linewidth}
  \captionsetup{type=figure,hypcap=false}
  \centering
  \includegraphics[width=0.95\linewidth]{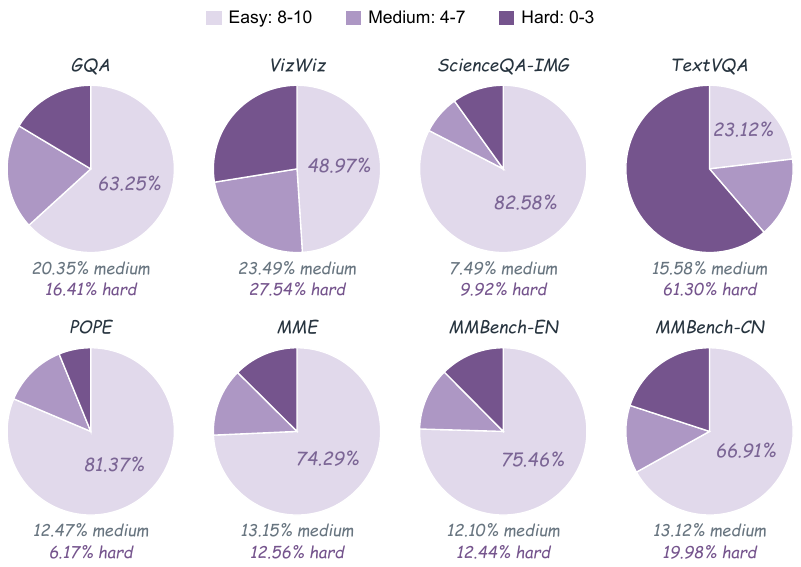}
  \caption{\textbf{Selection-difficulty distributions at 32 retained visual tokens.}
  Dataset order, colors, and annotations follow Figure~\ref{fig:difficulty-pies-128}.
  TextVQA is predominantly Hard (61.30\%), while ScienceQA-IMG and POPE
  remain predominantly Easy (82.58\% and 81.37\%, respectively).}
  \label{fig:difficulty-pies-32}
\end{minipage}

\clearpage
\subsection{Sensitivity to the Number of Random Trials}
\label{app:random-pass-trials}

Table~\ref{tab:random-pass-trial-overlap} compares ten random-pruning
trials with the cumulative set of 50 trials and with the independent
40 additional trials on GQA, TextVQA, and MMBench-EN at the same three
budgets. The independent comparison avoids sharing trials with the
ten-trial reference.

\noindent\begin{minipage}{\linewidth}
  \captionsetup{type=table,hypcap=false}
  \centering
  \caption{\textbf{Random-Pass overlap across trial counts (\%).}
  For each retained-token budget, ``10 vs.\ 50'' compares the original
  ten trials with all 50 trials; ``10 vs.\ 40'' compares them with the
  independent additional 40 trials. The overall values are unweighted
  means across the nine dataset--budget conditions.}
  \label{tab:random-pass-trial-overlap}
  \begingroup
  \fontsize{9}{10.8}\selectfont
  \setlength{\tabcolsep}{4pt}
  \renewcommand{\arraystretch}{1.13}
  \begin{tabular*}{\linewidth}{@{\extracolsep{\fill}}lrrrrrr@{}}
    \toprule
    & \multicolumn{2}{c}{128 tokens}
    & \multicolumn{2}{c}{64 tokens}
    & \multicolumn{2}{c}{32 tokens} \\
    \cmidrule(lr){2-3}\cmidrule(lr){4-5}\cmidrule(l){6-7}
    Dataset & 10 vs.\ 50 & \textbf{10 vs.\ 40}
            & 10 vs.\ 50 & \textbf{10 vs.\ 40}
            & 10 vs.\ 50 & \textbf{10 vs.\ 40} \\
    \midrule
    GQA        & 91.0 & 90.2 & 90.0 & 87.4 & 89.4 & 87.2 \\
    TextVQA    & 84.4 & 81.2 & 87.4 & 84.6 & 90.8 & 87.8 \\
    MMBench-EN & 97.2 & 96.8 & 95.2 & 94.2 & 93.6 & 92.4 \\
    \midrule
    \multicolumn{7}{@{}l@{}}{Overall: 10 vs.\ 50 = 91.0\%;\quad
      \textbf{10 vs.\ 40 = 89.1\%}} \\
    \bottomrule
  \end{tabular*}
  \endgroup
\end{minipage}

The mean overlap is 91.0\% with the cumulative 50 trials and 89.1\%
with the independent 40 trials. The independent overlap ranges from
81.2\% to 96.8\%, indicating that the observed stability varies across
datasets and budgets.
\clearpage

\clearpage
\end{document}